%% file: acl2025.tex
\pdfoutput=1
\documentclass[11pt]{article}
\usepackage{pifont}
\usepackage[final]{acl}

\usepackage{times}
\usepackage{latexsym}

\usepackage[T1]{fontenc}
\usepackage[utf8]{inputenc}

\usepackage{microtype}

\usepackage{inconsolata}

\usepackage{graphicx}

\usepackage{microtype}
\usepackage{graphicx}
\usepackage{booktabs} 

\usepackage{amsmath}
\usepackage{amssymb}
\usepackage{mathtools}
\usepackage{amsthm}

\usepackage{url}

\usepackage{colortbl}
\usepackage{multicol}
\usepackage{multirow}

\usepackage{fancybox}
\usepackage{listings}
\usepackage{times}
\usepackage{tabularx}
\usepackage{subcaption}
\usepackage{wrapfig}
\usepackage{algorithm}
\usepackage{algpseudocode}
\usepackage[most]{tcolorbox}
\usepackage{xltabular}
\usepackage{seqsplit}
\usepackage{amsmath,bm}
\usepackage{arydshln}
\usepackage{hyperref}

\definecolor{lightblue}{rgb}{0.68, 0.85, 0.9}
\definecolor{lightcoral}{rgb}{0.94, 0.5, 0.5}
\definecolor{kleinblue}{rgb}{0,0.18,0.65}
\definecolor{backblue}{RGB}{202, 225, 249}

\newcommand{\downarrownum}[1]{%
    \tikz[baseline=(text.base)]{
        \node[rectangle, rounded corners, fill=red!20, inner sep=1pt, scale=0.75] (text) {$\downarrow#1$};
    }%
}
\newcommand{\uparrownum}[1]{%
    \tikz[baseline=(text.base)]{
        \node[rectangle, rounded corners, fill=green!20, inner sep=1pt, scale=0.75] (text) {$\uparrow#1$};
    }%
}
\newcommand{\hkustgz}{\textsuperscript{\includegraphics[scale=0.02]{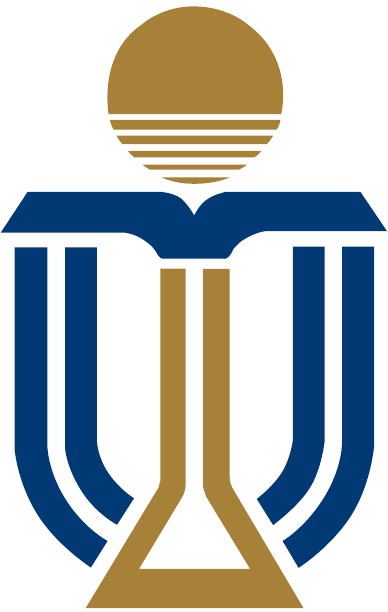}}}
\newcommand{\idea}{\textsuperscript{\includegraphics[scale=0.007]{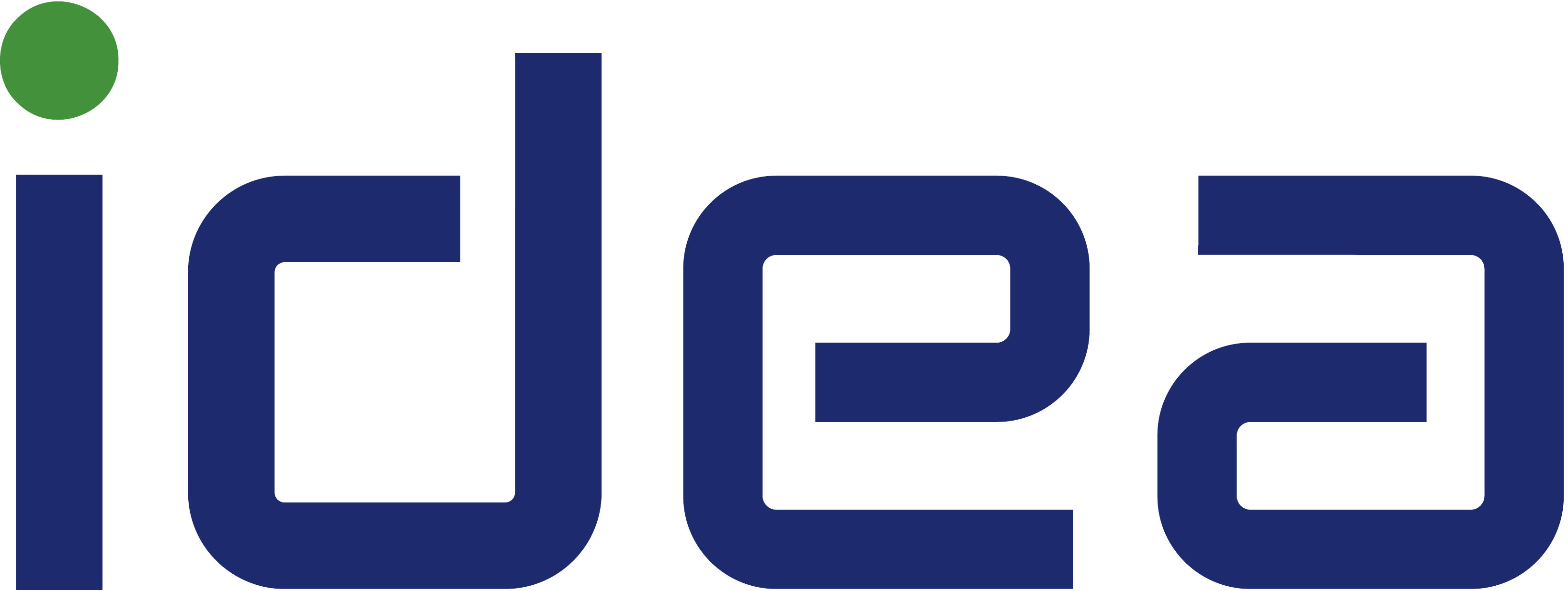}}}

\newtcbtheorem[number within=section]{exmp}{Example}%
{beforeafter skip balanced=.01in,colback=white!10!white,colframe=black!50!white,fonttitle=\bfseries, left=.02in, right=.02in,bottom=.02in, top=.02in}{exmp}
\newtcbtheorem[]{prompt}{Prompt}%
{colback=white!10!white,colframe=lightblue!50!white,fonttitle=\bfseries, left=.02in, right=.02in,bottom=.02in, top=.02in}{prompt}

\title{Parameter-Efficient Fine-Tuning via Circular Convolution}

\author{Aochuan Chen\hkustgz\thanks{Equal contribution.}~Jiashun Cheng\hkustgz\footnotemark[1]~Zijing Liu\idea~Ziqi Gao\hkustgz~Fugee Tsung\hkustgz~Yu Li\idea~Jia Li\hkustgz\thanks{Corresponding author.} \\
\hkustgz The Hong Kong University of Science and Technology (Guangzhou) \\
\idea International Digital Economy Academy \\
\href{mailto:jialee@hkust-gz.edu.cn}{\texttt{jialee@hkust-gz.edu.cn}}
}
\begin{document}
\maketitle

\input{tex/abstract}
\input{tex/introduction}

\input{tex/related_work}
\input{tex/method}
\input{tex/experiment}
\input{tex/conclusion}
\input{tex/limitation}
%\input{tex/acknowledgement}

\bibliography{anthology,acl2025}
% \bibliography{acl2025}

\clearpage
\appendix
\input{tex/appendix}

\end{document}

%% file: tex/abstract.tex
\begin{abstract}
Low-Rank Adaptation (LoRA) has gained popularity for fine-tuning large foundation models, leveraging low-rank matrices $\mathbf{A}$ and $\mathbf{B}$ to represent weight changes (\textit{i.e.,} $\Delta \mathbf{W} = \mathbf{B} \mathbf{A}$). This method reduces trainable parameters and mitigates heavy memory consumption associated with full delta matrices by sequentially multiplying $\mathbf{A}$ and $\mathbf{B}$ with the activation. Despite its success, the intrinsic low-rank characteristic may limit its performance. Although several variants have been proposed to address this issue, they often overlook the crucial computational and memory efficiency brought by LoRA. In this paper, we propose \underline{C}ir\underline{c}ular \underline{C}onvolution \underline{A}daptation (C$^3$A), which not only achieves high-rank adaptation with enhanced performance but also excels in both computational power and memory utilization. Extensive experiments demonstrate that C$^3$A consistently outperforms LoRA and its variants across various fine-tuning tasks. Our code is available at \href{https://github.com/huggingface/peft}{Hugging Face PEFT}.
\end{abstract}

%% file: tex/introduction.tex
\section{Introduction}

\begin{figure}
    \begin{center}   \includegraphics[width=1.\linewidth]{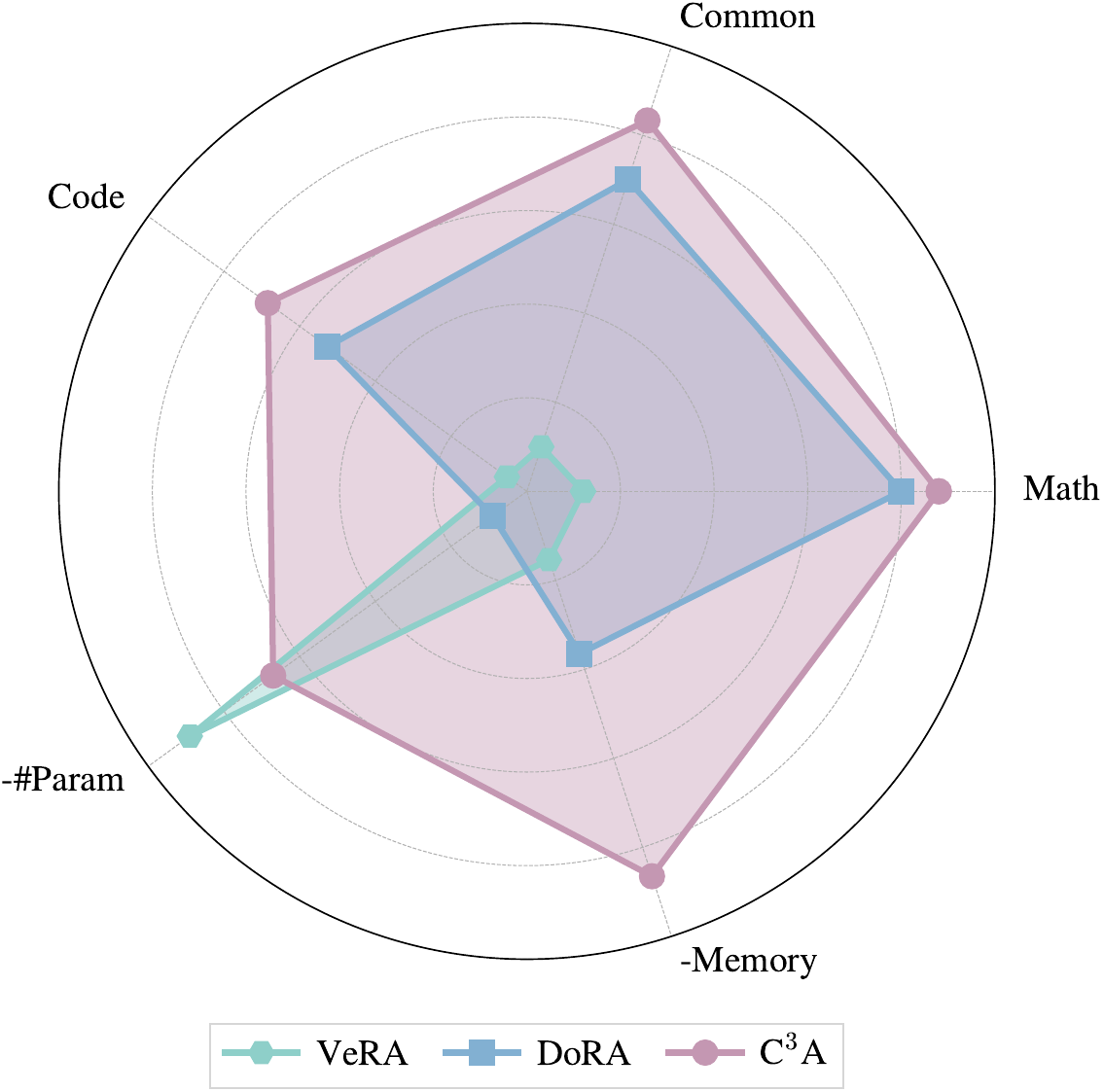}
    \end{center}
    \caption{Performance comparison of C$^3$A and other methods relative to LoRA on LLaMA-8B. Higher values are better across all metrics. See Table \ref{tab:common} and Table \ref{tab:math_code} for more statistics.}
    \label{fig:teaser}
    \vspace{-5mm}
\end{figure}

In recent years, Large Foundation Models (LFMs) have witnessed a pronounced ascendance in both scholarly and practical realms, attributable to their exceptional efficacy across diverse tasks in natural language processing (NLP) \citep{brown2020language, touvron2023llama}, computer vision (CV) \citep{radford2021learning, kirillov2023segment}, and other domains \citep{li2024zerog, li2024glbench,li2025g}. Distinguished by an extensive parameter count and significant computational requisites, these models have established unprecedented benchmarks in both accuracy and versatility. Nonetheless, their considerable size and intricate structure present formidable obstacles for efficient fine-tuning, especially within resource-constrained environments \citep{malladi2023fine, zhang2024adam}. To mitigate these challenges, parameter-efficient fine-tuning (PEFT) techniques \citep{peft}, exemplified by Low-Rank Adaptation (LoRA) \citep{hu2021lora}, have emerged as highly effective solutions.

LoRA reduces the number of trainable parameters by leveraging low-rank matrices to approximate alterations in weights, thereby facilitating fine-tuning without degrading the model's efficacy. Specifically, LoRA can be articulated mathematically as follows:
\begin{align*}
    \mathbf{Wx} = (\mathbf{W}_0 + \Delta \mathbf{W})\mathbf{x} = \mathbf{W}_0\mathbf{x} + \mathbf{B(Ax)},
\end{align*}
where $\mathbf{W},\mathbf{W}_0,\Delta \mathbf{W}\in \mathbb{R}^{d_1\times d_2}$ are weight matrices, $\mathbf{B}\in\mathbb{R}^{d_1\times r}, \mathbf{A}\in\mathbb{R}^{r\times d_2}$ are low-rank matrices formulated to construct $\Delta\mathbf{W}$, and $\mathbf{x}\in\mathbb{R}^{d_2}$ are the activations. The number of trainable parameters is \( r(d_1+d_2) \), thereby motivating the selection of \( r \ll \min(d_1,d_2) \) (\textit{e.g.,} \( r = 8 \) for \( d_1 = d_2 = 1024 \)) to attain elevated parameter efficiency. Nonetheless, as elaborated by \citet{zeng2023expressive}, the potential of LoRA to encapsulate a target model is inherently constrained by \( r \). In an effort to reconcile the dichotomy between performance and efficiency, \citet{kopiczko2023vera} introduced Vector Random Matrix Adaptation (VeRA). VeRA attains comparable performance with a markedly reduced count of trainable parameters via fixed random-matrix projections. However, despite its minimal parameter count, VeRA demands considerable computational resources and memory capacity due to the extensive nature of the random matrices employed for projection. As depicted in Figure \ref{fig:teaser}, other representative works share the same resource problem. This precipitates the following open research question within the scope of PEFT:

\begin{center}
\textit{Beyond low parameter counts, how to achieve high-rank adaptation without incurring significant costs of time and memory?}
\end{center}

To address this question, we introduce \underline{C}ir\underline{c}ular \underline{C}onvolution \underline{A}daptation (C$^3$A), which incorporates the circular convolution operator \citep{bamieh2018discovering}. Circular convolution has garnered significant attention in both signal processing \citep{Li2020RevisitingLC} and cryptography \citep{117146} due to its exceptional efficiency and compactness. This operator can be equivalently expressed as multiplication by a circulant matrix, providing rank flexibility that is independent of the number of trainable parameters. Furthermore, by employing the Fast Fourier Transform (FFT), C$^3$A achieves superior time and memory efficiency compared to the direct multiplication \citep{bamieh2018discovering}, which makes it competitive with LoRA in terms of efficiency. 

In addition, as explicated by \citet{dosovitskiy2020image}, dense linear layers exhibit a deficiency of inductive biases, engendering a complex optimization landscape. Consequently, this hampers the effectiveness of transformers in comparison to Convolutional Neural Networks (CNNs) under conditions of limited data availability. Within the framework of a constrained training dataset for the downstream task, we postulate that a robust inductive bias could potentially augment adaption performance. The circular pattern in C$^3$A serves precisely as such an inductive bias.

In summary, circular convolution presents a promising solution for circumventing the rank limitations of LoRA at minimal costs. Our contributions can be summarized as follows:

\ding{182} We introduce C$^3$A, a novel approach for PEFT. This method leverages the circular convolution operation and its equivalent circulant matrix to provide a flexible rank, which is free of linear constraint by the number of trainable parameters.

\ding{183} Leveraging the elegant diagonalization of the circulant matrix, we implement both the forward pass and backpropagation using FFT. With the incorporation of FFT, the computation and memory efficiency of C$^3$A excels. C$^3$A strikes a unique balance between performance an efficiency. 

\ding{184} To offer greater flexibility in controlling the number of trainable parameters, we extend C$^3$A by incorporating block-circular convolution, which results in block-circulant matrices. This extension allows C$^3$A to achieve fully customizable parameter counts as well as adaptable rank configurations.

\ding{185} We validate C$^3$A through comprehensive fine-tuning experiments across diverse tasks, which demonstrate C$^3$A's outstanding accuracy and memory merits compared to existing methods.

%% file: tex/related_work.tex
\section{Related Work}

\subsection{Parameter-Efficient Fine-Tuning}

Research on PEFT has generally progressed along three main directions. The first direction involves partially updating the pre-trained neural network (\textit{e.g.,} the layer norm \citep{basu2024strong} or the biases \citep{zaken2021bitfit}). Traditional methods relied on hand-crafted heuristics \citep{raghu2019rapid} to identify which parameters are crucial and should be fine-tuned. More advanced approaches employ optimization techniques \citep{guo2020parameter, xu2021raise, fu2023effectiveness}. For example, \citet{guo2020parameter} reformulated such a discrete optimization problem into a continuous one by employing Bernoulli masks and the Gumbel-softmax approximation \citep{jang2016categorical}.

The second direction emerged to maintain the integrity of the pre-trained model while enabling a high degree of parameter sharing through adapter-based methods \citep{he2021towards, rebuffi2017learning, ruckle2020adapterdrop, liu2022few, lian2022scaling}. These works focus on integrating additional modules, termed adapters, to fit the downstream task, effectively decoupling the pre-trained model parameters from those specific to the downstream task. Prompt Tuning \citep{brown2020language, gao2020making, chen2023understanding, zhang2024visual} and Prefix Tuning \citep{li2021prefix, jia2022visual} also fall into this category, despite ignoring potential semantic meanings.

The final direction is characterized by delta-weight-based methods, such as Low-Rank Adaptation (LoRA) \citep{hu2021lora} and Orthogonal Fine-tuning (OFT) \citep{qiu2023controlling}. These methods bridge the gap between the pre-trained model and the downstream task by adaptive delta weights, which are stored separately while used in combination with the pre-trained weights. This unique design enables disentanglement of the pretrained and downstream-specific weights. Namely, it achieves parameter sharing and preserves the ability to integrate without additional inference cost. LoRA models the delta-weights by an additive matrix while OFT does it by a multiplicative one. To further improve either parameter efficiency or performance, many variants has been proposed for both of the methods \citep{kopiczko2023vera, liu2024dora, liu2023parameter, yuan2024bridging, hayou2024lora+}. However, these methods can hardly achieve high parameter efficiency and performance without incurring heavy computation and memory usage. 

\subsection{Circular Convolution}

Circular convolution has been extensively studied in signal processing \citep{4309918, McGillem1984ContinuousAD, Li2020RevisitingLC} and cryptography \citep{117146, gong2024practical}. Owing to its computational advantages, circular convolution has also been explored in machine learning for generating long embeddings of high-dimensional data \citep{yu2014circulant} and compressing heavily parameterized layers \citep{cheng2015exploration, 10.1145/3123939.3124552}. Remarkably, it achieves these efficiencies without significant performance degradation, which makes it a promising technique for fine-tuning applications.

Despite its success in small neural networks such as LeNet \citep{cheng2015exploration}, circular convolution has not demonstrated lossless performance in modern large foundational models (LFMs) or even in their base architecture, the transformer. This limitation may be attributed to the conflict between its high intrinsic bias (\textit{i.e.,} the circulant pattern) and the vast amount of data required for training LFMs. Conversely, when fine-tuning LFMs, it is often impractical to collect as much data as needed for training from scratch. In such scenarios, the intrinsic bias of circular convolution could potentially serve as a regularization mechanism, thereby benefiting the optimization process of fine-tuning.

%% file: tex/method.tex
\section{Method}

\begin{figure*}[htb]
\centering
\includegraphics[width=1.0\textwidth]{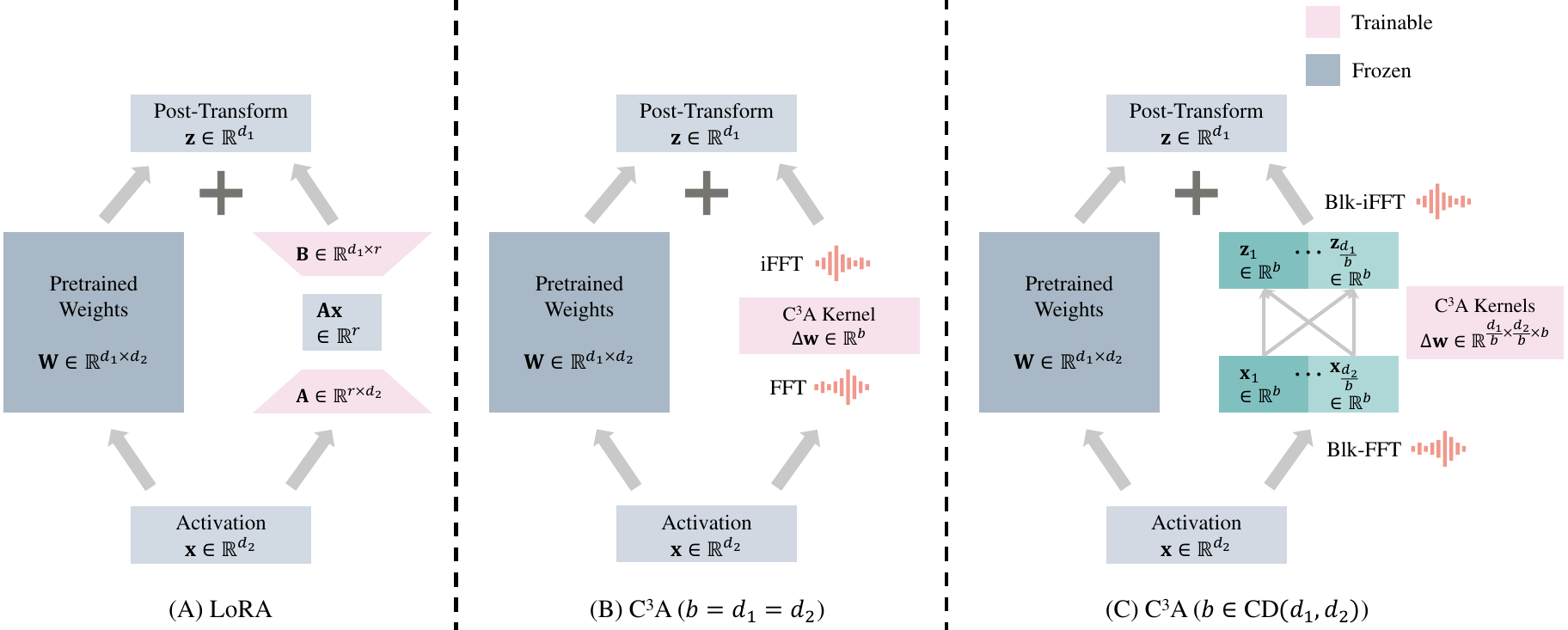} 
\caption{\textbf{Overview of LoRA (A) and our C$^3$A (B,C) method.} In LoRA, only low-rank matrices $\mathbf{A}$ and $\mathbf{B}$ are trained and the delta weight is represented by their product (\textit{i.e.,} $\Delta \mathbf{W}=\mathbf{BA}$). The total trainable parameter number is $r(d_1+d_2)$, which is assosiated with the rank of the delta weight. In C$^3$A, circular convolution kernels $\Delta\mathbf{w}$ are tuned to adapt to the downstream task and the delta weight is represented by the (block-)circular matrix they construct (\textit{i.e.,} $\Delta\mathbf{W}=\mathcal{C}_{\mathrm{(blk)}}(\Delta\mathbf{w})$). The total trainable parameter count is $\frac{d_1d_2}{b}$, which disentangles with the rank of the delta weight. Here, $b$ is the block size and it should be a common divisor (CD) of $d_1$ and $d_2$.} 
\label{fig:main}
\end{figure*}

In this section, we present C$^3$A (see an overview in Figure \ref{fig:main}), a novel PEFT approach based on the circular convolution. C$^3$A follows LoRA's setting of learning an additive linear operation over the original dense linear transformation. However, instead of using low-rank decomposition and the matrix multiplication operator, C$^3$A resorts to circular convolution as this additive linear operation.

\subsection{Notations}
\label{sec:method-notation}
The adapted weight matrix, the original weight matrix, and the delta matrix are denoted by $\mathbf{W}$, $\mathbf{W}_0$, and $\Delta\mathbf{W}$, respectively ($\mathbf{W},\mathbf{W}_0,\Delta \mathbf{W}\in \mathbb{R}^{d_1\times d_2}$). The activation vector of the previous layer is denoted by $\mathbf{x} \in \mathbb{R}^{d_2} $. The post-transformation vector is $\mathbf{z}$, where $\mathbf{z} = \mathbf{W}\mathbf{x}\in \mathbb{R}^{d_1}$, and the incremental part is denoted by $\Delta \mathbf{z}$, where $\Delta \mathbf{z} = \Delta\mathbf{W}\mathbf{x} \in \mathbb{R}^{d_1}$. The matrices $\mathbf{A}$ and $\mathbf{B}$ are low-rank matrices introduced by LoRA to represent $\Delta \mathbf{W}$, with $r$ being their rank. $r_v$ specifies the rank of the random projection matrix used in VeRA. The circular convolution kernel of C$^3$A is denoted by $\Delta\mathbf{w}$ and the circular convolution operator by $\star$. The loss function is represented by $\mathcal{L}$. The Fast Fourier Transform and its inverse are denoted by $\mathrm{FFT}$ and $\mathrm{iFFT}$. The Hadamard product is denoted by $\circ$.

\subsection{Circular Convolution}
\label{sec:method-convop}
Firstly, for simplicity, we assume $d_1=d_2=d$ and $\Delta \mathbf{w}\in \mathbb{R}^d$. The circular convolution operator is defined as $\Delta\mathbf{z} = \Delta\mathbf{w} \star \mathbf{x} = \mathcal{C}(\Delta\mathbf{w})\mathbf{x}$, where $\mathcal{C}(\cdot)$ is a function which takes a vector and outputs the corresponding circulant matrix. Concretely, the first row of $\mathcal{C}(\Delta\mathbf{w})$ is $\Delta\mathbf{w}$ and the following rows are equal to the row above them periodically shifted to the right by one element. In math,

{\small{\begin{align*}
    \mathcal{C}(\Delta\mathbf{w}) = 
    \begin{bmatrix}
    \Delta w_1 & \Delta w_2 & \cdots & \Delta w_{d-1} & \Delta w_d \\
    \Delta w_d & \Delta w_1 & \cdots & \Delta w_{d-2} & \Delta w_{d-1} \\
    \cdots & \cdots & \cdots & \cdots & \cdots\\
    \Delta w_3 & \Delta w_4 & \cdots & \Delta w_1 & \Delta w_2 \\
    \Delta w_2 & \Delta w_3 & \cdots & \Delta w_d & \Delta w_1 \\
    \end{bmatrix}.
\end{align*}}}%

Theoretically, the rank of $\mathcal{C}(\Delta\mathbf{w})$ is given by $d - \mathrm{Deg}(\mathrm{gcd}(f(x), x^d - 1))$ \citep{s1-31.4.445}, where $\mathrm{Deg}(\cdot)$ denotes the degree of a polynomial, $f(x)$ is the polynomial associated with $\Delta\mathbf{w}$ (\textit{i.e.,} $f(x) = \sum_{i=1}^d \Delta w_{i} x^{i-1}$), and $\mathrm{gcd}(\cdot)$ represents the greatest common divisor. Consequently, the theoretical upper bound on the rank of $\mathcal{C}(\Delta\mathbf{w})$ is $d$. By learning $\Delta\mathbf{w}$ in the $\mathbb{R}^n$ oracle, C$^3$A automatically achieves dynamic rank selection, which is not linearly constrained by the number of learnable parameters, unlike LoRA.

To achieve high efficiency, enlightened by \citet{10.1145/3123939.3124552}, we leverage the beautiful circulant structure of $\mathcal{C}(\Delta\mathbf{w})$, which makes it diagonalizable by the Fourier basis ($\mathbf{F}$) . In math, it can be described as $\mathcal{C}(\Delta \mathbf{w}) = \mathbf{F} \frac{\Lambda}{d} \mathbf{F}^{-1}$ \citep{10.5555/248979}, where $\Lambda$ is its eigenvalues and can be calculated by a Fourier transform of the first row (\textit{i.e.,} $\Lambda = \mathrm{diag}(\mathbf{F}\Delta\mathbf{w})$). Therefore, we can calculate $\Delta \mathbf{w} \star \mathbf{x}$ as 

\vspace{-3mm}
{\small{\begin{align}
    \begin{split}
    \label{eq:circ_conv}
    \Delta\mathbf{w} \star \mathbf{x} &= \mathbf{F}\mathrm{diag}(\frac{\mathbf{F}\Delta\mathbf{w}}{d})\mathbf{F}^{-1}\mathbf{x} \\
    &= \mathrm{FFT}(\mathrm{FFT}(\Delta\mathbf{w})\circ\mathrm{iFFT}(\mathbf{x})).
    \end{split}
\end{align}}}% 

\subsection{Backpropagation}
\label{sec:method-backprop}  
To effectuate backpropagation with optimal efficiency, it is imperative to obtain the analytical derivatives of the loss function \(\mathcal{L}\) with respect to \(\Delta\mathbf{w}\) and \(\mathbf{x}\). Following the approach outlined in \citet{10.1145/3123939.3124552}, we aim to explain backpropagation using simpler language. By applying the chain rule, these derivatives are delineated as follows:

\vspace{-3mm}
{\small{\begin{align}
    \label{eq:chainrule}
    \frac{\partial \mathcal{L}}{\partial \mathbf{x}} = \frac{\partial \Delta\mathbf{z}}{\partial \mathbf{x}} \frac{\partial \mathcal{L}}{\partial \Delta\mathbf{z}}, \qquad \frac{\partial \mathcal{L}}{\partial \Delta\mathbf{w}} = \frac{\partial \Delta\mathbf{z}}{\partial \Delta\mathbf{w}} \frac{\partial \mathcal{L}}{\partial \Delta\mathbf{z}}.
\end{align}}}%

Given that \(\Delta\mathbf{z} = \mathcal{C}(\Delta\mathbf{w}) \mathbf{x}\), it logically follows that \(\frac{\partial \Delta\mathbf{z}}{\partial \mathbf{x}} = \mathcal{C}(\Delta\mathbf{w})\). Concerning \(\frac{\partial \Delta\mathbf{z}}{\partial \Delta\mathbf{w}}\), we observe the commutative property of the circular convolution operation (\textit{i.e.,} \(\mathcal{C}(\Delta\mathbf{w}) \mathbf{x} = \mathcal{C}(\mathbf{x}) \Delta\mathbf{w}\)), which implies \(\frac{\partial \Delta\mathbf{z}}{\partial \Delta\mathbf{w}} = \mathcal{C}(\mathbf{x})\). Substituting these findings into Equation \ref{eq:chainrule}, we derive:

\vspace{-3mm}
{\small{\begin{align*}
    \frac{\partial \mathcal{L}}{\partial \mathbf{x}} = \mathcal{C}(\Delta\mathbf{w}) \frac{\partial \mathcal{L}}{\partial \Delta\mathbf{z}}, \qquad \frac{\partial \mathcal{L}}{\partial \Delta\mathbf{w}} = \mathcal{C}(\mathbf{x}) \frac{\partial \mathcal{L}}{\partial \Delta\mathbf{z}}.
\end{align*}}}%

These expressions can also be interpreted as circular convolutions:

\vspace{-3mm}
{\small{\begin{align*}
    \frac{\partial \mathcal{L}}{\partial \mathbf{x}} = \Delta\mathbf{w} \star \frac{\partial \mathcal{L}}{\partial \Delta\mathbf{z}}, \qquad \frac{\partial \mathcal{L}}{\partial \Delta\mathbf{w}} = \mathbf{x} \star \frac{\partial \mathcal{L}}{\partial \Delta\mathbf{z}}.
\end{align*}}}%

By meticulously executing this derivative computation in accordance with Equation \ref{eq:circ_conv}, backpropagation can harness the computational efficacy facilitated by the FFT algorithm.

\subsection{Block-Circular Convolution}
\label{sec:method-blockconv}
Notwithstanding the elegance and efficiency of the circular convolution operator, it is subject to two fundamental limitations stemming from the constraint that the convolution kernel must match the dimensions of the activation vector: \ding{172} \textit{It is inapplicable to non-square weight matrices.} \ding{173} \textit{The count of learnable parameters remains fixed.} The first restriction hampers its applicability in scenarios such as fine-tuning a LLaMA3-8B model, where the weight matrix dimensions include $4096 \times 1024$. The second constraint diminishes the adaptability of C$^3$A, presenting challenges in addressing complex downstream tasks that necessitate a greater number of learnable parameters. To mitigate these limitations, we employ block-circular convolution \citep{10.1145/3123939.3124552}. By partitioning the activation vector $\mathbf{x}$ and the post-transformation vector $\Delta\mathbf{z}$ into blocks of identical size, unique convolution kernels can be allocated to each pair of these blocks. Specifically,

\vspace{-3mm}
{\small{\begin{align*}
    \mathbf{x} &= \begin{bmatrix} \mathbf{x}_1 & \mathbf{x}_2 & \cdots & \mathbf{x}_\frac{d_2}{b} \end{bmatrix} \\ \Delta\mathbf{z} &= \begin{bmatrix} \Delta\mathbf{z}_1 & \Delta\mathbf{z}_2 & \cdots & \Delta\mathbf{z}_\frac{d_1}{b} \end{bmatrix},
\end{align*}}}%

where $b$ is the block size and $b$ need to be a common divisor of $d_1$ and $d_2$. We will need $\frac{d_1d_2}{b^2}$ convolution kernels to densely connect these blocks, which can be expressed in math as 

\vspace{-3mm}
{\small{\begin{align*}
    \Delta\mathbf{z}_i = \sum_{j=1}^{\frac{d_2}{b}}\Delta\mathbf{w}_{ij} \star \mathbf{x}_j, i\in \{1,2,\cdots,\frac{d_1}{b}\}.
\end{align*}}}%

This calculation can be represented by a block-circular matrix:

\vspace{-3mm}
{\small{\begin{align}
    \Delta\mathbf{z} &= \mathcal{C}_{\mathrm{blk}}(\Delta\mathbf{w})\mathbf{x} \\
    \mathcal{C}_{\mathrm{blk}}(\Delta\mathbf{w}) &=
    \begin{bmatrix} 
    \mathcal{C}(\Delta\mathbf{w}_{11}) & \cdots & \mathcal{C}(\Delta\mathbf{w}_{1\frac{d_2}{b}}) \\
    \mathcal{C}(\Delta\mathbf{w}_{21}) & \cdots & \mathcal{C}(\Delta\mathbf{w}_{2\frac{d_2}{b}}) \\
    \cdots & \cdots & \cdots \\
    \mathcal{C}(\Delta\mathbf{w}_{\frac{d_1}{b}1}) & \cdots & \mathcal{C}(\Delta\mathbf{w}_{\frac{d_1}{b}\frac{d_2}{b}}) 
    \end{bmatrix}. \label{eq:blk-circulant-matrix}
\end{align}}}%

We refer our readers to Algorithm \ref{alg:blk-circular-conv} in Appendix \ref{sec:appendix-implementations} for a Pytorch implementation. In this context, $\Delta\mathbf{w}_{ij} \in \mathbb{R}^{b}$, and it follows that $\frac{d_1d_2}{b^2} b = \frac{d_1d_2}{b}$ represents the number of learnable parameters. Notably, the parameter $b$ serves as a hyperparameter modulating the quantity of learnable parameters, analogous to the role of $r$ in LoRA. It is imperative to distinguish, however, that whereas $r$ simultaneously governs the rank of the delta matrix and the number of learnable parameters, $b$ exclusively influences the latter. This disentanglement of matrix rank and parameter count facilitates greater adaptability and potentially yields superior outcomes. 

\subsection{Complexity Analysis}
\label{sec:method-complexity}
We compare the time complexity and space complexity of LoRA, VeRA and C$^3$A in Table \ref{tab:complexity}. Detailed analysis follows in this section.

\subsubsection{Time Complexity}
LoRA integrates low-rank matrices $\mathbf{A}$ and $\mathbf{B}$, which are successively multiplied with the activation vector, resulting in a computational complexity of $\mathcal{O}(r(d_1 + d_2))$. Generally, $r \ll \min(d_1, d_2)$. In contrast, VeRA, despite its high-rank structure and relatively few trainable parameters, suffers from a prohibitive computational complexity of $\mathcal{O}(r_v(d_1 + d_2))$, where $r_v$ can exceed $\max(d_1, d_2)$. Consequently, striking an optimal balance between high rank and computational efficiency remains an elusive task.

On GPUs, the cuFFT backend automatically parallelizes FFT operations along the axes not being transformed, with the degree of parallelism $p$ determined by the available resources. Thanks to the $\mathcal{O}(n \log n)$ complexity of the FFT algorithm used in Equation \ref{eq:circ_conv}, C$^3$A achieves a time complexity of $\mathcal{O}(\frac{(d_1 + d_2)}{p}\log{b} + \frac{d_1 d_2}{b})$. The first term is the time complexity for FFT and the second term is for aggregation. In practical scenarios, $b$ is chosen as the greatest common divisor of $d_1$ and $d_2$ to achieve a high compression ratio. Given that, C$^3$A is comparable to LoRA in time complexity.

\input{tables/complexity}

\subsubsection{Space Complexity}
We analyze the space complexity of LoRA, VeRA, and C$^3$A during training. The differences among these methods primarily arise from the trainable parameters and the auxiliary tensors required for the forward pass and backpropagation. LoRA does not rely on auxiliary tensors, while VeRA necessitates 2 random projection matrices, with a total size of $r_v(d_1 + d_2)$. Since $r_v$ is by no means negligible, the memory usage of VeRA is significantly larger than that of LoRA.

In terms of C$^3$A, the only additional auxiliary tensor would be of size $pb \leq \min(d_1, d_2)$, which is reserved by the FFT algorithm. By selecting an appropriate $b$, which is often close to the greatest common divisor of $d_1$ and $d_2$, the space complexity of C$^3$A is optimized. Furthermore, because $p$ scales with the available resources, the algorithm inherently manages dynamic memory consumption without additional effort.

%% file: tables/complexity.tex
% \begin{wraptable}{r}{100mm}
\begin{table}[htb]
\centering
\resizebox{\linewidth}{!}{
    \begin{tabular}{lccc}
    \toprule
    \multirow{2}{*}{Method} & \multirow{2}{*}{Time} & \multicolumn{2}{c}{Space} \\
    \cmidrule(lr){3-4}
    & & \# Param & \# Other \\
    \midrule
    LoRA & $\mathcal{O}(r(d_1+d_2))$ & $r(d_1+d_2)$ & 0 \\
    VeRA & $\mathcal{O}(r_v(d_1+d_2))$ & $r_v+d_1$ & $r_v(d_1+d_2)$ \\
    C$^3$A & $\mathcal{O}(\frac{d_1+d_2}{p}\log{b}+\frac{d_1d_2}{b})$ & $\frac{d_1d_2}{b}$ & $pb$ \\
    \bottomrule
    \end{tabular}
}
\caption{Time and space complexity comparison of LoRA, VeRA and C$^3$A. We split the space complexity into Parameter number and Other auxiliary tensors to help better understand the differences. We highlight that in practice, to achieve similar performance, $\frac{\max(d_1,d_2)}{b} \leq r \ll r_v$.}
\label{tab:complexity}
\end{table}

%% file: tex/experiment.tex
\section{Experiment}

\input{tables/glue}

\paragraph{Baselines.}
We compare our C$^3$A with several representative PEFT methods, including BitFit \citep{zaken2021bitfit}, (IA)$^3$ \citep{liu2022few}, LoRA \citep{hu2021lora}, VeRA \citep{kopiczko2023vera}, BOFT \citep{liu2023parameter} and DoRA \citep{liu2024dora}.
BitFit fine-tunes biases. (IA)$^3$ is a SOTA method adding adapters. LoRA uses low-rank decomposition to compress additive delta matrices. VeRA reduces LoRA's trainable parameters while maintaining a high rank. BOFT compresses multiplicative delta matrices via orthogonal decomposition and butterfly factorization. DoRA separately learns magnitude and direction of delta matrices.

\input{tables/commonsense}
\input{tables/math_code}

\subsection{GLUE Benchmark}

\paragraph{Settings.}  
We assess our C$^3$A on the GLUE benchmark \citep{wang2018glue} covering various tasks like single-sentence classification, similarity, paraphrase, and inference. See Table \ref{tab:data-glue} in Appendix \ref{sec:appendix-glue} for more details. Datasets are split into train, validation, and test sets. Models are chosen based on validation performance and evaluated on the test set. We fine-tune RoBERTa-Base and RoBERTa-Large models \citep{liu2019roberta}. Unique hyperparameters are from original papers (\textit{e.g.,} for VeRA's $r$ and BOFT's $b$ and $m$). Count of trainable parameters (\# Params) exclude the classification head, which is uniform across methods. Shared hyperparameters (\textit{i.e.,} learning rates for the classification head and other parameters) are determined by hyperparameter search. Regarding the memory test, input data is fixed to 256 tokens with a batch size of 64 for consistency.

\paragraph{Results.}
Results are presented in Table \ref{tab:glue}. Overall, C$^3$A$_{b=768/1}$ and C$^3$A$_{b=1024/1}$ achieve superior or comparable performance to baseline methods, despite using an exceptionally small number of trainable parameters. As the number of trainable parameters increases, models like C$^3$A$_{b=768/6}$ and C$^3$A$_{b=1024/8}$ significantly outperform the baselines. Moreover, compared to (IA)$^3$, LoRA, VeRA, and BOFT, C$^3$A distinguishes itself with remarkable memory efficiency. The only method with better memory efficiency is BitFit, which serves as an upper bound since it does not introduce new parameters. Furthermore, most of the delta matrices identified by C$^3$A are of full rank, indicating maximum capacity \citep{zeng2023expressive} and providing a theoretical basis for outstanding performance.

\subsection{Instruction Fine-tuning}

\paragraph{Commonsense Reasoning.}
Our training utilizes Commonsense170K \citep{hu2023llm}, aggregating multiple-choice questions from BoolQ \citep{clark2019boolq}, PIQA \citep{bisk2020piqa}, SIQA \citep{sap2019socialiqa}, HellaSwag \citep{zellers2019hellaswag}, WinoGrande \citep{Sakaguchi2021}, ARC-e, ARC-c \citep{Clark2018}, and OBQA \citep{Mihaylov2018}. Evaluation follows \citet{hu2023llm} and \citet{liu2024dora}, using greedy search and first-keyword appearance for answer determination.

\paragraph{Mathematical Reasoning.}
We use MetaMathQA \citep{yu2023metamath}, containing 395K QA pairs from GSM8K \citep{cobbe2021training} and MATH \citep{hendrycks2020measuring}. Following \citet{yu2023metamath}, we evaluate using greedy search on test sets, requiring chain-of-thought reasoning \citep{Wei2022}.

\paragraph{Code Generation.}
Training uses Magicoder-Evol-Instruct-110k \citep{Wei2024}, a decontaminated subset of WizardCoder \citep{Luo2024}. Evaluation on HumanEval \citep{Chen2021}, MBPP \citep{Austin2021}, and their Plus variants follows EvalPlus \citep{Liu2024}, reporting Pass@1 metrics.

\paragraph{Results.} In Table \ref{tab:common} and Table \ref{tab:math_code}, our principal experimental observations are summarized. The C$^3$A framework consistently surpasses LoRA and other baselines within the LLaMA series, with particular efficacy demonstrated in the most recent model, LLaMA3-8B. Noteworthy is the significant enhancement in the efficacy of LLaMA3-8B as a foundational model following the implementation of more sophisticated post-training techniques. This underscores the criticality of optimizing the fine-tuning protocols for this advanced model. It is also remarkable that C$^3$A achieves such results while employing less than half the parameter count of LoRA. Furthermore, C$^3$A achieves high rank adaptation and delivers optimal performance while maintaining remarkably low memory consumption. In contrast, VeRA and DoRA require substantially greater memory resources, and BoFT results in an OOM condition for a single H800, thereby underscoring the practicality and efficiency of C$^3$A. Taken together, the findings robustly underscore the superior efficacy of the C$^3$A methodology. We refer readers to Appendix \ref{sec:appendix-inst-ft-examples} for examples of model answers after different tuning methods.

\subsection{Ablation Study}

\begin{figure}[htb]
    \centering
    \includegraphics[width=\linewidth]{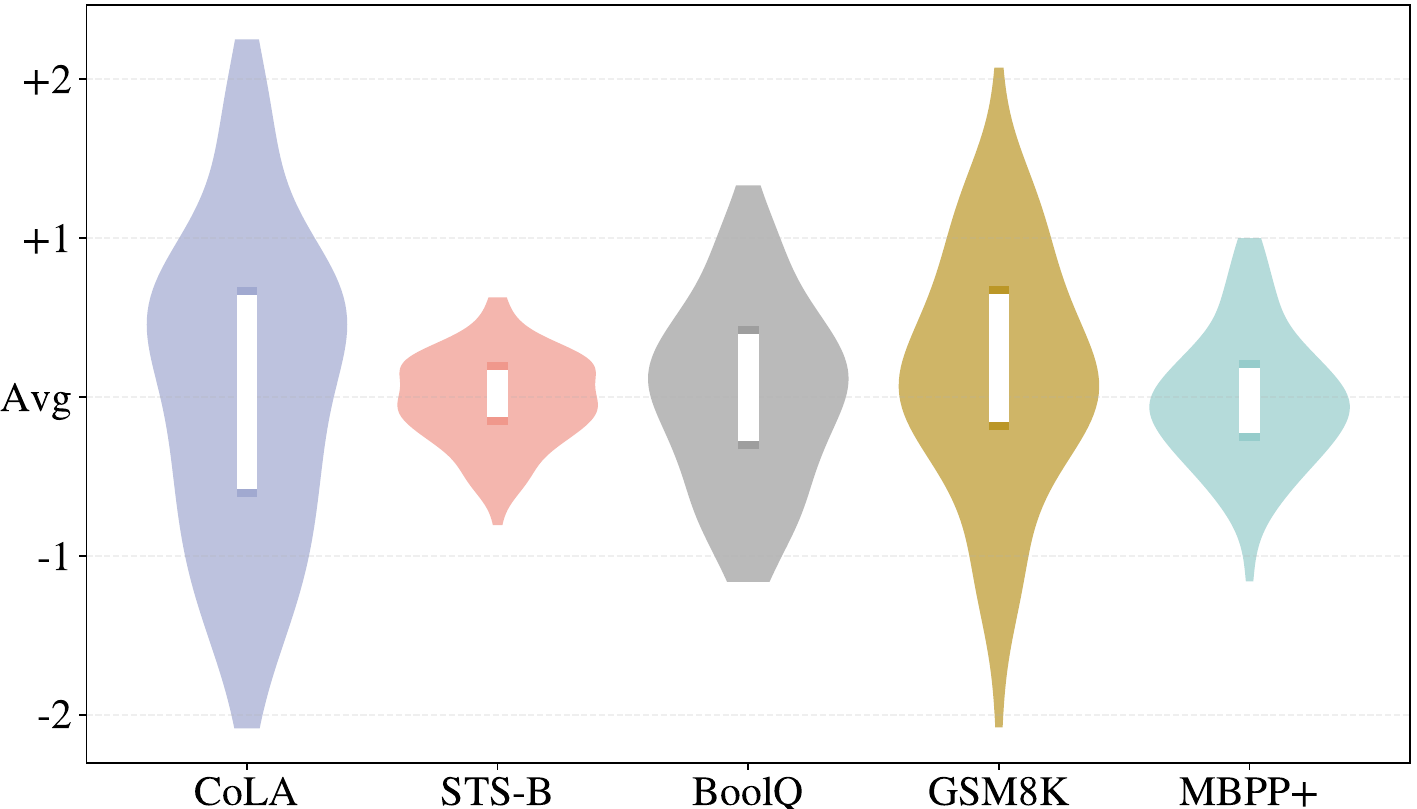}
    \caption{Violin plots for runs on different tasks. Within each violin plot, a white bar indicates the range of average performance for each initialization strategy. It is clear that the choice of initialization does not affect the final result more than the intrinsic stochasticity.}
    \label{fig:ablation-initialization}
\end{figure}

\paragraph{Initialization.}
We investigated initialization effects in C$^3$A compared to LoRA, which is known for initialization sensitivity due to its $\mathbf{A}$ and $\mathbf{B}$ matrices \citep{hayou2024impact}. We evaluated multiple initialization strategies across five distinct tasks. Each initialization method was tested with 5 independent runs per task, producing a total of 20 runs per task. As shown in Figure \ref{fig:ablation-initialization}, testing zero, Gaussian, Kaiming uniform, and Xavier uniform initializations for C$^3$A's convolution kernels revealed performance variations mostly within standard deviations, demonstrating C$^3$A's robustness to initialization choices.

\paragraph{Expressiveness.} 
We demonstrate C$^3$A's expressiveness on a synthetic dataset by placing 8 cluster centers on a 2D plane and sampling 30 points from their Gaussian distributions. A 3-layer MLP is used to classify these clusters. To compare the expressiveness of LoRA and C$^3$A, we replace the middle layer with either a low-rank layer or a circulant layer, ensuring that both layers have the same number of trainable parameters for a fair comparison. 

\begin{figure}[htb]
  \centering
  \includegraphics[width=\linewidth]{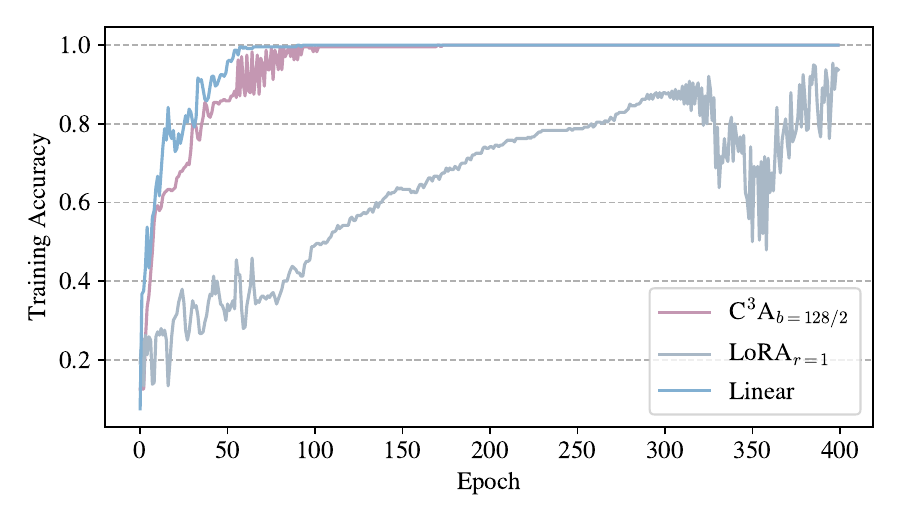}
  \caption{Training curves on the synthetic dataset. We refer our readers to Figure \ref{fig:synthetic-points} in Appendix \ref{sec:appendix-synthetic-dataset} for a visualization of the synthetic dataset.}
  \label{fig:expressive_training_curve}
\end{figure}

The results are presented in Figure \ref{fig:expressive_training_curve}. We observe that LoRA$_{r=1}$ struggles with this simple classification task. In contrast, C$^3$A$_{b=128/2}$, despite using the same number of parameters, achieves a perfect classification, comparable to a standard linear layer. This demonstrates the high expressiveness of C$^3$A given the same parameter budget.

\paragraph{Scaling.}

\begin{figure}
    \centering
    \includegraphics[width=\linewidth]{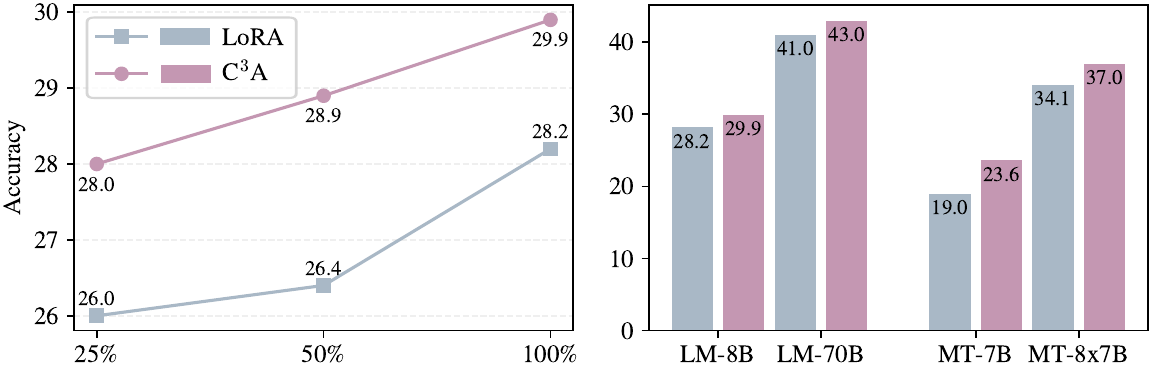}
    \caption{Data and model scaling of C$^3$A compared to LoRA. LM specifies LLaMA3 and MT specifies Mistral.}
    \label{fig:ablation-scaling}
\end{figure}

To investigate the scaling effect of C$^3$A, we examine it from both data and model perspectives, following the setup in \citet{he2025rasa}. From a data perspective, we focus on the MATH dataset and vary the training data used. As shown in Figure \ref{fig:ablation-scaling}, we observe that C$^3$A's performance improves faster than LoRA as more data is added, suggesting it is more effective at leveraging additional data. From the model perspective, we evaluate C$^3$A in small models (LLaMA3-8B, Mistral-7B) and large models (LLaMA3-70B, Mistral-8x7B). Results show that C$^3$A outperforms the LoRA baseline in both settings, indicating that its benefits generalize to different model scales. Experiments on various data quantities and model sizes showcase C$^3$A's scalability and robustness, making it a promising approach for adapting language models.

% \paragraph{Impact of Placement in Transformers.}

%% file: tables/glue.tex
\begin{table*}[htb]
\centering
\resizebox{1.\textwidth}{!}{
\begin{tabular}{llrrccccccc}
\toprule
& Methods & \# Params & Mem & SST-2 & MRPC & CoLA & QNLI & RTE & STS-B & Avg. \\
\midrule
\parbox[t]{4mm}{\multirow{8}{*}{\rotatebox[origin=c]{90}{\textsc{Base}}}} 
& Full & 124M & 17.19G & 94.01\textsubscript{$\pm$0.39} & \textbf{87.10}\textsubscript{$\pm$0.79} & 62.00\textsubscript{$\pm$1.16} & 92.40\textsubscript{$\pm$0.28} & \textbf{77.33}\textsubscript{$\pm$2.68} & \textbf{90.70}\textsubscript{$\pm$0.14} & \textbf{83.92}\\
& BitFit & 0.102M & 12.60G & 93.30\textsubscript{$\pm$0.30} & 85.80\textsubscript{$\pm$0.21} & 59.21\textsubscript{$\pm$1.74} & 91.96\textsubscript{$\pm$0.18} & 73.07\textsubscript{$\pm$1.34} & 90.18\textsubscript{$\pm$0.17} & 82.25 \\
& (IA)$^3$ & 0.111M & 19.86G & 92.98\textsubscript{$\pm$0.34} & 85.86\textsubscript{$\pm$0.59} & 60.49\textsubscript{$\pm$1.09} & 91.56\textsubscript{$\pm$0.17} & 69.10\textsubscript{$\pm$1.18} & 90.06\textsubscript{$\pm$0.21} & 81.67 \\
& LoRA$_{r=8}$ & 0.295M & 13.75G & \textbf{94.50}\textsubscript{$\pm$0.41} & 85.68\textsubscript{$\pm$0.74} & 60.95\textsubscript{$\pm$1.57} & \textbf{92.54}\textsubscript{$\pm$0.20} & 76.68\textsubscript{$\pm$1.42} & 89.76\textsubscript{$\pm$0.39} & 83.35 \\
& VeRA$_{r=1024}$ & 0.043M & 15.51G & 93.97\textsubscript{$\pm$0.17} & 86.23\textsubscript{$\pm$0.41} & 62.24\textsubscript{$\pm$1.91} & 91.85\textsubscript{$\pm$0.17} & 75.74\textsubscript{$\pm$1.56} & 90.27\textsubscript{$\pm$0.25} & 83.38 \\
& BOFT$^{m=2}_{b=8}$ & 0.166M & 14.11G & 93.23\textsubscript{$\pm$0.50} & 84.37\textsubscript{$\pm$0.54} & 59.50\textsubscript{$\pm$1.25} & 91.69\textsubscript{$\pm$0.12} & 74.22\textsubscript{$\pm$0.84} & 89.63\textsubscript{$\pm$0.37} & 82.11 \\
& C$^3$A$_{b=768/1}$ & 0.018M & 12.83G & 93.42\textsubscript{$\pm$0.26} & 86.33\textsubscript{$\pm$0.32} & 61.83\textsubscript{$\pm$0.96} & 91.83\textsubscript{$\pm$0.04} & 76.17\textsubscript{$\pm$1.39} & 90.46\textsubscript{$\pm$0.29} & 83.34 \\
& C$^3$A$_{b=768/6}$ & 0.111M & 12.72G & 94.20\textsubscript{$\pm$0.16} & 86.67\textsubscript{$\pm$0.54} & \textbf{62.48}\textsubscript{$\pm$1.20} & 92.32\textsubscript{$\pm$0.25} & 77.18\textsubscript{$\pm$1.41} & 90.16\textsubscript{$\pm$0.42} & 83.84 \\

\midrule[0.6pt]
\midrule
\parbox[t]{4mm}{\multirow{8}{*}{\rotatebox[origin=c]{90}{\textsc{Large}}}} 
& Full & 354M & 43.40G & 95.75\textsubscript{$\pm$0.45} & \textbf{88.35}\textsubscript{$\pm$0.64} & 64.87\textsubscript{$\pm$1.25} & 92.40\textsubscript{$\pm$0.28} & 84.48\textsubscript{$\pm$1.14} & 91.65\textsubscript{$\pm$0.14} & 86.25\\
& BitFit & 0.271M & 30.65G & 95.09\textsubscript{$\pm$0.27} & 88.10\textsubscript{$\pm$0.76} & 65.40\textsubscript{$\pm$0.76} & 94.06\textsubscript{$\pm$0.14} & 82.60\textsubscript{$\pm$1.15} & 91.73\textsubscript{$\pm$0.20} & 86.16 \\
& (IA)$^3$ & 0.295M & 48.81G & 95.32\textsubscript{$\pm$0.20} & 87.06\textsubscript{$\pm$0.57} & 66.52\textsubscript{$\pm$1.10} & 94.18\textsubscript{$\pm$0.15} & 84.33\textsubscript{$\pm$2.38} & 91.58\textsubscript{$\pm$0.39} & 86.50 \\
& LoRA$_{r=8}$ & 0.786M & 34.12G & 95.53\textsubscript{$\pm$0.35} & 86.12\textsubscript{$\pm$0.86} & 65.16\textsubscript{$\pm$0.76} & 93.73\textsubscript{$\pm$0.30} & 83.75\textsubscript{$\pm$0.51} & 91.46\textsubscript{$\pm$0.21} & 85.96 \\
& VeRA$_{r=256}$ & 0.061M & 34.16G & \textbf{95.83}\textsubscript{$\pm$0.43} & 87.72\textsubscript{$\pm$0.55} & 63.66\textsubscript{$\pm$1.45} & 94.11\textsubscript{$\pm$0.20} & 83.03\textsubscript{$\pm$1.65} & 91.12\textsubscript{$\pm$0.37} & 85.91 \\
& BOFT$^{m=2}_{b=8}$ & 0.442M & 34.98G & 95.76\textsubscript{$\pm$0.41} & 88.28\textsubscript{$\pm$0.33} & 64.72\textsubscript{$\pm$2.37} & 93.89\textsubscript{$\pm$0.14} & 82.82\textsubscript{$\pm$1.40} & 91.03\textsubscript{$\pm$0.32} & 86.08 \\
& C$^3$A$_{b=1024/1}$ & 0.049M & 31.83G & 95.78\textsubscript{$\pm$0.05} & 88.02\textsubscript{$\pm$0.62} & 66.59\textsubscript{$\pm$1.20} & 94.22\textsubscript{$\pm$0.25} & 82.89\textsubscript{$\pm$0.67} & \textbf{91.86}\textsubscript{$\pm$0.14} & 86.56 \\
& C$^3$A$_{b=1024/8}$ & 0.393M & 31.79G & 95.78\textsubscript{$\pm$0.15} & 88.09\textsubscript{$\pm$0.47} & \textbf{67.18}\textsubscript{$\pm$1.92} & \textbf{94.26}\textsubscript{$\pm$0.19} & \textbf{84.62}\textsubscript{$\pm$1.36} & 91.81\textsubscript{$\pm$0.36} & \textbf{86.96} \\
\bottomrule
\end{tabular}
}
\caption{Performance of different PEFT methods on the GLUE benchmark. We fine-tune pre-trained RoBERTa-Base and -Large models on 6 datasets. We report the Matthew's Correlation Coefficient (MCC) for CoLA, Pearson Correlation Coefficient (PCC) for STS-B, and accuracy (Acc.) for all the remaining tasks. For each metric, a higher score indicates better performance. ``Avg.'' denotes the average score of each method across all datasets. The best results for each dataset are highlighted in \textbf{bold}. \# Params does not include the classification head since each method uses a head of the same size. Memory cost (Mem) is measured on fixed length (\textit{i.e.,} $256$) data with a batchsize of $64$.}
\label{tab:glue}
\end{table*}

%% file: tables/commonsense.tex
\begin{table*}[htb]
    \centering
    \resizebox{1.\textwidth}{!}{
        \begin{tabular}{llrrrrrrrrrrr}
            \toprule
            & Methods & Params (\%) & Mem & BoolQ & PIQA & SIQA & HellaS.&  WinoG. & ARC-e & ARC-c & OBQA & Avg. \\
            \midrule \specialrule{0em}{1.5pt}{1.5pt}
            & ChatGPT & - & - & 73.1 & 85.4 & 68.5 & 78.5 & 66.1 & 89.8 & 79.9 & 74.8 & 77.0 \\
            
            \midrule[0.6pt]
            \midrule[0.6pt]

            % \multicolumn{1}{l}{\textbf{\textit{LLaMA2$_{\textsc{7b}}$}}} \\
            \parbox[t]{4mm}{\multirow{5}{*}{\rotatebox[origin=c]{90}{LLaMA2-7B}}} & LoRA$_{r=32}$ & 0.83 & 41.71G & 71.0 & 81.4 & 79.6 & 87.4 & 83.2 & 82.6 & 67.5 & 81.5 & 79.3 \\
            \cmidrule(lr){2-13}
            % & VeRA$_{r=16384}$ & 0.05 & 57.88G & 70.9\downarrownum{0.1} & 83.0\uparrownum{1.6} & 79.4\downarrownum{0.2} & \textbf{89.5}\uparrownum{2.1} & 83.3\uparrownum{0.1} & 83.1\uparrownum{0.5} & 64.8\downarrownum{2.7} & 81.9\uparrownum{0.4} & 79.5\uparrownum{0.2} \\
            & VeRA$_{r=16384}$ & 0.05 & 57.88G & 68.9\downarrownum{2.1} & 81.0\downarrownum{0.4} & 77.4\downarrownum{2.2} & 87.5\uparrownum{0.1} & 81.3\downarrownum{1.9} & 81.1\downarrownum{1.5} & 62.8\downarrownum{4.7} & 79.9\downarrownum{1.6} & 77.5\downarrownum{1.8} \\
            & BOFT$^{m=2}_{b=8}$ & \multicolumn{11}{c}{Out of Memory} \\
            & DoRA$_{r=32}$ & 0.84 & 55.24G & 72.3\uparrownum{1.3} & \textbf{84.2}\uparrownum{2.8} & 78.7\downarrownum{0.9} & \textbf{88.5}\uparrownum{1.1} & 84.0\uparrownum{0.8} & 80.7\downarrownum{1.9} & \textbf{69.1}\uparrownum{1.6} & \textbf{83.7}\uparrownum{2.2} & 80.2\uparrownum{0.9} \\
            & C$^3$A$_{b=4096/32}$ & 0.35 & 44.89G & \textbf{73.4}\uparrownum{2.4} & 83.4\uparrownum{2.0} & \textbf{82.1}\uparrownum{2.5} & 88.3\uparrownum{0.9} & \textbf{84.9}\uparrownum{1.7} & \textbf{86.6}\uparrownum{4.0} & 66.7\downarrownum{0.8} & 82.5\uparrownum{1.0} & \textbf{81.0}\uparrownum{1.7} \\
            
            \midrule[0.6pt]
            \midrule[0.6pt]
            
            % \multicolumn{1}{l}{\textbf{\textit{LLaMA3$_{\textsc{8b}}$}}} \\
            \parbox[t]{4mm}{\multirow{5}{*}{\rotatebox[origin=c]{90}{LLaMA3-8B}}} & LoRA$_{r=32}$ & 0.70 & 51.18G & 73.8 & 88.2 & 80.4 & 94.0 & 85.5 & 87.5 & 78.1 & 84.0 & 83.9 \\
            \cmidrule(lr){2-13}
            % & VeRA$_{r=16384}$ & 0.04 & 66.03G & 73.4\downarrownum{0.4} & 88.7\uparrownum{0.5} & 81.2\uparrownum{0.8} & 95.5\uparrownum{1.5} & 85.1\downarrownum{0.4} & 85.7\downarrownum{1.8} & 76.8\downarrownum{1.3} & 83.6\downarrownum{0.4} & 83.7\downarrownum{0.2} \\
            & VeRA$_{r=16384}$ & 0.04 & 66.03G & 72.2\downarrownum{1.6} & 87.5\downarrownum{0.7} & 80.0\downarrownum{0.4} & 94.3\uparrownum{0.3} & 83.9\downarrownum{1.6} & 84.5\downarrownum{3.0} & 75.6\downarrownum{2.5} & 82.4\downarrownum{1.6} & 82.5\downarrownum{1.4} \\
            & BOFT$^{m=2}_{b=8}$ & \multicolumn{11}{c}{Out of Memory} \\
            & DoRA$_{r=32}$ & 0.71 & 63.37G & 75.3\uparrownum{1.5} & 88.9\uparrownum{0.7} & \textbf{82.2}\uparrownum{1.8} & \textbf{96.7}\uparrownum{2.7} & 86.0\uparrownum{0.5} & 89.0\uparrownum{1.5} & \textbf{79.8}\uparrownum{1.7} & 84.9\uparrownum{0.9} & 85.3\uparrownum{1.4} \\
            & C$^3$A$_{b=4096/32}$ & 0.26 & 56.08G & \textbf{76.9}\uparrownum{3.1} & \textbf{91.4}\uparrownum{3.2} & 82.1\uparrownum{1.7} & 94.9\uparrownum{0.9} & \textbf{86.9}\uparrownum{1.4} & \textbf{89.6}\uparrownum{2.1} & 79.4\uparrownum{1.3} & \textbf{86.1}\uparrownum{2.1} & \textbf{85.9}\uparrownum{2.0} \\
            \bottomrule
        \end{tabular}
    }
    \caption{Comparison of various methods on the LLaMA2-7B and LLaMA3-8B models across eight commonsense reasoning datasets. For each language model, the best result on each dataset is shown in \textbf{bold}. The symbols $\uparrow$ and $\downarrow$ indicate relative improvement and decrease, respectively, compared to the LoRA method. Experiments are conducted on a single H800 GPU with 80GB HBM.
    }
    \label{tab:common}
\end{table*}

%% file: tables/math_code.tex
\begin{table*}[htb]
    \centering
    \resizebox{1.\textwidth}{!}{
        \begin{tabular}{llrrrrrrrrr}
            \toprule
            & \multirow{2}{*}{Methods} & \multirow{2}{*}{Params (\%)} & \multicolumn{2}{c}{Math} & \multirow{2}{*}{Avg.} & \multicolumn{4}{c}{Code} & \multirow{2}{*}{Avg.} \\
            \cmidrule(lr){4-5} \cmidrule(lr){7-10}
            & & & GSM8K & MATH & & HumanEval & HumanEval+ & MBPP & MBPP+ & \\
            
            \midrule \specialrule{0em}{1.5pt}{1.5pt}
            % \multicolumn{1}{l}{\textbf{\textit{LLaMA2$_{\textsc{7b}}$}}} \\
            \parbox[t]{4mm}{\multirow{5}{*}{\rotatebox[origin=c]{90}{LLaMA2-7B}}} & Zero-Shot & - & 7.3 & 1.1 & 4.2 & 11.0 & 9.8 & 30.2 & 24.1 & 18.8 \\
            & LoRA$_{r=32}$ & 0.83 & 60.5 & 11.7 & 36.1 & 31.7 & 28.0 & 35.4 & 30.4 & 31.4 \\
            \cmidrule(lr){2-11}
            % & VeRA$_{r=16384}$ & 0.05 & 61.9\uparrownum{1.4} & 10.0\downarrownum{1.7} & 36.0\downarrownum{0.1} & 30.4\downarrownum{1.3} & 26.1\downarrownum{1.9} & 36.6\uparrownum{1.2} & 31.1\uparrownum{0.7} & 31.0\downarrownum{0.3} \\
            & VeRA$_{r=16384}$ & 0.05 & 58.5\downarrownum{2.0} & 9.4\downarrownum{2.3} & 34.0\downarrownum{2.1} & 28.9\downarrownum{2.8} & 24.6\downarrownum{3.4} & 35.1\downarrownum{0.3} & 29.6\downarrownum{0.8} & 29.5\downarrownum{1.9} \\
            % & BOFT$^{m=2}_{b=8}$ &  & 60.3\downarrownum{0.2} & 10.6\downarrownum{1.1} & 35.5\downarrownum{0.6} & 32.0\uparrownum{0.3} & 28.7\uparrownum{0.7} & 34.2\downarrownum{1.2} & 30.1\downarrownum{0.3} & 31.3\downarrownum{0.1} \\
            & DoRA$_{r=32}$ & 0.84 & \textbf{62.3}\uparrownum{1.8} & 12.4\uparrownum{0.7} & \textbf{37.3}\uparrownum{1.2} & 33.2\uparrownum{1.5} & 28.8\uparrownum{0.8} & \textbf{36.8}\uparrownum{1.4} & 31.8\uparrownum{1.4} & 32.7\uparrownum{1.3} \\
            & C$^3$A$_{b=4096/32}$ & 0.35 & 61.5\uparrownum{1.0} & \textbf{13.0}\uparrownum{1.3} & 37.2\uparrownum{1.1} & \textbf{33.8}\uparrownum{2.1} & \textbf{29.0}\uparrownum{1.0} & 36.5\uparrownum{1.1} & \textbf{31.9}\uparrownum{1.5} & \textbf{32.8}\uparrownum{1.4} \\
            
            \midrule[0.6pt]
            \midrule[0.6pt]
            
            % \multicolumn{1}{l}{\textbf{\textit{LLaMA3$_{\textsc{8b}}$}}} \\
            \parbox[t]{4mm}{\multirow{5}{*}{\rotatebox[origin=c]{90}{LLaMA3-8B}}} & Zero-Shot & - & 33.1 & 5.3 & 19.2 & 33.5 & 29.3 & 61.4 & 51.6 & 44.0 \\
            & LoRA$_{r=32}$ & 0.70 & 77.2 & 28.2 & 52.7 & 57.9 & 52.4 & 64.8 & 55.3 & 57.6 \\
            \cmidrule(lr){2-11}
            % & VeRA$_{r=16384}$ & 0.04 & 76.0\downarrownum{1.2} & 28.5\uparrownum{0.3} & 52.2\downarrownum{0.5} & 57.8\downarrownum{0.1} & 52.3\downarrownum{0.1} & 65.5\uparrownum{0.7} & 55.1\downarrownum{0.2} & 57.7\uparrownum{0.1} \\
            & VeRA$_{r=16384}$ & 0.04 & 76.0\downarrownum{1.2} & 26.7\downarrownum{1.5} & 51.4\downarrownum{1.3} & 56.3\downarrownum{1.6} & 50.8\downarrownum{1.6} & 64.0\downarrownum{0.8} & 53.6\downarrownum{1.7} & 56.2\downarrownum{1.4} \\
            % & BOFT$^{m=2}_{b=8}$ &  & 76.9\downarrownum{0.3} & 28.0\downarrownum{0.2} & 52.4\downarrownum{0.3} & 58.4\uparrownum{0.5} & 51.2\downarrownum{1.2} & 64.9\uparrownum{0.1} & 55.6\uparrownum{0.3} & 57.5\downarrownum{0.1} \\
            & DoRA$_{r=32}$ & 0.71 & 77.2\uparrownum{0.0} & \textbf{30.0}\uparrownum{1.8} & 53.6\uparrownum{0.9} & 59.1\uparrownum{1.2} & 52.9\uparrownum{0.5} & \textbf{65.7}\uparrownum{0.9} & 55.8\uparrownum{0.5} & 58.4\uparrownum{0.8} \\
            & C$^3$A$_{b=4096/32}$ & 0.26 & \textbf{78.4}\uparrownum{1.2} & 29.9\uparrownum{1.7} & \textbf{54.1}\uparrownum{1.4} &  \textbf{59.5}\uparrownum{1.6} & \textbf{53.8}\uparrownum{1.4} & 65.4\uparrownum{0.6} & \textbf{56.1}\uparrownum{0.8} & \textbf{58.7}\uparrownum{1.1} \\
            \bottomrule
        \end{tabular}
    }
    \caption{Comparison of various methods on the LLaMA2-7B and LLaMA3-8B models across math reasoning and code generation datasets. For each language model, the best result on each dataset is shown in \textbf{bold}. The symbols $\uparrow$ and $\downarrow$ indicate relative improvement and decrease, respectively, compared to the LoRA method. Experiments are conducted on a single H800 GPU with 80GB HBM.}
    \label{tab:math_code}
\end{table*}

%% file: tex/conclusion.tex
\section{Conclusion}

This manuscript introduces C$^3$A, a novel PEFT method. Unlike LoRA's low-rank decomposition, C$^3$A utilizes circular convolution and circulant matrices to construct the delta weight matrix. This approach allows independent control over the delta weight matrix's rank and the number of trainable parameters, enabling high-rank adaptation with limited parameter size. By employing FFT in forward and backward propagation, C$^3$A achieves significant computational and memory efficiency, presenting a compelling alternative to LoRA for model fine-tuning.

%% file: tex/limitation.tex
\section{Limitations}

% Despite C$^3$A's efficiency and effectiveness within limited parameter budgets, it has some drawbacks. First, the requirement of $b\in \mathrm{gcd} (d_1,d_2)$ limits its flexibility for all transformers. Second, the FFT integration requires kernel support, which is more developed for NVIDIA GPUs than edge devices.

Despite C$^3$A's efficiency and effectiveness within limited parameter budgets, it has some intrinsic drawbacks. First, due to the requirement of $b\in \mathrm{gcd} (d_1,d_2)$, C$^3$A may not be flexible enough to tackle all transformer architectures. 
Besides, the integration of FFT in C$^3$A necessitates specialized kernel support for optimal performance. While this support is well-developed and readily available for NVIDIA GPUs, it may not be as mature or optimized for edge devices with different hardware architectures. This can potentially hinder the deployment and efficiency of C$^3$A on resource-constrained edge platforms.

%% file: tex/appendix.tex
% \onecolumn

\section*{Appendix}

\setcounter{section}{0}
\setcounter{figure}{0}
\setcounter{table}{0}
\setcounter{algorithm}{0}

\renewcommand{\thefigure}{A\arabic{figure}}
\renewcommand{\theHfigure}{A\arabic{figure}}
\renewcommand{\thetable}{A\arabic{table}}
\renewcommand{\theHtable}{A\arabic{table}}
\renewcommand{\thealgorithm}{A\arabic{algorithm}}
\renewcommand{\theHalgorithm}{A\arabic{algorithm}}

\definecolor{exmp_inst}{HTML}{82B0D2}
\definecolor{exmp_c3a}{HTML}{C4A5DE}
\definecolor{exmp_lora}{HTML}{A1A9D0}

\section{Implementations}\label{sec:appendix-implementations}

\input{algs/block_circular_convolution}

We present the PyTorch implementation of Block-Circular Convolution in Algorithm \ref{alg:blk-circular-conv}. Furthermore, due to the inefficiency of directly assigning entries (as shown in Equation \ref{eq:blk-circulant-matrix}), we derive an alternative algorithm to compute the $\Delta\mathbf{W}$ more efficiently. Rather than direct assignment, we employ a forward process on the Identity matrix. Mathematically, this can be expressed as

\vspace{-3mm}
{\small{\begin{align*}
    \Delta\mathbf{W} &= \mathcal{C}_{\mathrm{blk}}(\Delta\mathbf{w}) \\
    &= \mathcal{C}_{\mathrm{blk}}(\Delta\mathbf{w}) \cdot \mathbf{I}_{d_2} \\
    &= \mathcal{C}_{\mathrm{blk}}(\Delta\mathbf{w}) \cdot \left [ \mathbf{e}_1, \mathbf{e}_2, \cdots, \mathbf{e}_{d_2} \right ] \\
    &= \left [ \mathcal{C}_{\mathrm{blk}}(\Delta\mathbf{w})\mathbf{e}_1, \mathcal{C}_{\mathrm{blk}}(\Delta\mathbf{w})\mathbf{e}_2, \cdots, \mathcal{C}_{\mathrm{blk}}(\Delta\mathbf{w})\mathbf{e}_{d_2} \right ]\\
    &= \left [ \Delta\mathbf{w} \star \mathbf{e}_1,\Delta\mathbf{w} \star \mathbf{e}_2, \cdots, \Delta\mathbf{w} \star \mathbf{e}_{d_2} \right ].
\end{align*}}}%

Where $\mathbf{I}_{d_2} \in \mathbb{R}^{d_2\times d_2}$ represents an Identity matrix and $\mathbf{e}_i$ is the $i$th column of it. In pytorch, we can efficiently compute the iFFT of $\{\mathbf{e}_i\}_{i=1,2,\cdots,d_2}$ by a column-wise iFFT of $\mathbf{I}_{d_2}$. We present the Pytorch implementation in Algorithm \ref{alg:fast-delta} as well.

\input{algs/fast_delta}

\section{Image Classification}

\paragraph{Settings.}
In this study, we concentrate on the task of image classification leveraging Vision Transformer (ViT) models. Specifically, we employ both the Base and Large variants of this prominent foundational computer vision model, as delineated by \citep{dosovitskiy2020image}. These ViT models undergo pre-training on the expansive ImageNet-21K dataset \citep{ridnik2021imagenet}. During the fine-tuning phase, we use an eclectic array of datasets encompassing Pets \citep{pets}, Cars \citep{cars}, DTD \citep{dtd}, EuroSAT \citep{euro}, FGVC \citep{fgvc}, and RESISC \citep{resisc}. Statistics for these datasets are provided in Table \ref{tab:data-cv}.

\input{tables/vit_dataset_detail}

\input{tables/image_classification}

\paragraph{Results.} 
Table \ref{tab:image_classification} delineates a comprehensive summary of the outcomes derived from six distinct image classification datasets employing the ViT Base and Large models. The LoRA and C$^3$A techniques exhibit significant enhancements in performance relative to Head Tuning, thereby underscoring their efficacy within the realm of image classification. Remarkably, our methodology demonstrates a performance on par with LoRA while necessitating only half of the parameter count.

\section{GLUE Benchmark Details}\label{sec:appendix-glue}

\input{tables/glue_details}

The General Language Understanding Evaluation (GLUE) benchmark is a collection of nine natural language understanding tasks designed to test a model's performance on a diverse set of natural language understanding challenges. It was introduced by researchers from New York University, the University of Washington, and DeepMind in a 2018 paper titled ``GLUE: A Multi-Task Benchmark and Analysis Platform for Natural Language Understanding'' \citep{wang2018glue}.

The nine tasks included in the benchmark are:

\begin{enumerate}
    \item \textbf{CoLA (Corpus of Linguistic Acceptability):} A binary classification task that tests a model's ability to determine whether a given sentence is grammatically acceptable.
    \item \textbf{SST-2 (Stanford Sentiment Treebank):} A binary sentiment classification task using movie review excerpts.
    \item \textbf{MRPC (Microsoft Research Paraphrase Corpus):} A binary classification task to determine whether two sentences are semantically equivalent.
    \item \textbf{STS-B (Semantic Textual Similarity Benchmark):} A regression task that scores the semantic similarity of sentence pairs on a scale from 1 to 5.
    \item \textbf{QQP (Quora Question Pairs):} A binary classification task to determine whether two questions are semantically equivalent.
    \item \textbf{MNLI (Multi-Genre Natural Language Inference):} A ternary classification task that tests a model's ability to perform textual entailment across multiple genres.
    \item \textbf{QNLI (Question Natural Language Inference):} A binary classification task that tests a model's ability to determine whether a sentence contains the answer to a given question.
    \item \textbf{RTE (Recognizing Textual Entailment):} A binary classification task that tests a model's ability to determine whether a premise entails a hypothesis.
    \item \textbf{WNLI (Winograd Natural Language Inference):} A binary classification task that tests a model's ability to resolve ambiguous pronouns in a sentence.
\end{enumerate}

The GLUE benchmark has become a widely-used standard for evaluating the performance of pre-trained language models and fine-tuned models on a variety of natural language understanding tasks. The benchmark provides a single-number metric, the GLUE score, which is an average of the scores on each individual task. This allows for easy comparison of different models and architectures. We refer readers to Table \ref{tab:data-glue} for more details.

% The introduction of GLUE has spurred significant advancements in natural language understanding, with models like BERT, RoBERTa, and ALBERT achieving human-level performance on some tasks. However, the benchmark has also faced criticism for its limitations, such as the lack of more challenging tasks and the potential for models to exploit superficial patterns in the data.

\section{Instruction Fine-tuning Examples}
\label{sec:appendix-inst-ft-examples}

\begin{exmp}{MATH}{mathexample}
\textit{Below is an instruction that describes a task.
Write a response that appropriately completes the request.} 

\textbf{Instruction:} \textcolor{exmp_inst}{Find the number of square units bounded by the $x$-axis, the $y$-axis and the line $x+ y= 6$. Answer the above question. First think step by step and then answer the final number.}

\textbf{LoRA Answer:} \textcolor{exmp_lora}{The line $x+y=6$ intersects the $x$-axis at $x=6$ and the $y$-axis at $y=6$. So, the line divides the square with side length 6 into two triangles. The area of each triangle is $\frac{1}{2} \times 6 \times 6=18$. So, the area of the square is $2 \times 18=\boxed{36}$. The final answer is: 36.}

\textbf{C$^3$A Answer:} \textcolor{exmp_c3a}{The line $x+y=6$ intersects the $x$-axis at $(6,0)$ and the $y$-axis at $(0,6)$. Thus, the line divides the first quadrant into two triangles, one with base $6$ and height $6$ and the other with base $6$ and height $0$. The area of the first triangle is $(6 \times 6) / 2 = 18$ square units, and the area of the second triangle is $(6 \times 0)/2=0$ square units. Thus, the total area is $18+0=\boxed{18}$ square units. The final answer is: 18.}
\end{exmp}

\begin{exmp}{GSM8K}{gsm8kexample}
\textit{Below is an instruction that describes a task.
Write a response that appropriately completes the request.} 

\textbf{Instruction:} \textcolor{exmp_inst}{    Tim makes a special honey and jam mix. To make 1 jar of the mixture it takes 2 pounds of passion fruit and 1.5 pounds of jackfruit. It also takes 2 pounds of special honey. The passion fruit costs \$6 per pound, the jackfruit is \$8 per pound, and the honey is \$10 per pound.  He sells each jar for \$50.  How much profit does he make selling 10 jars? Answer the above question. First think step by step and then answer the final number.}

\textbf{LoRA Answer:} \textcolor{exmp_lora}{The passion fruit costs $6 \times 2=12$. The jackfruit costs $8 \times 1.5= 12$. The honey costs $10 \times 2 = 20$. So the total cost is $ 12+12+20 = 44$. So he makes a profit of $50 \times 10 - 44=\boxed{460}$. The final answer is: 460.}

\textbf{C$^3$A Answer:} \textcolor{exmp_c3a}{The passion fruit costs $6 \times 2 = 12$. The jackfruit costs $8 \times 1.5 = 12$. So the fruit costs $12+12=24$. The honey costs $10 \times 2=20$. So the total cost per jar is $24+20=44$. So he makes a profit of $50-44=6$ per jar. So he makes a profit of $6 \times 10=\boxed{60}$. The final answer is: 60.}
\end{exmp}

\section{Synthetic Dataset}\label{sec:appendix-synthetic-dataset}

\begin{figure}
    \centering
    \includegraphics[width=\linewidth]{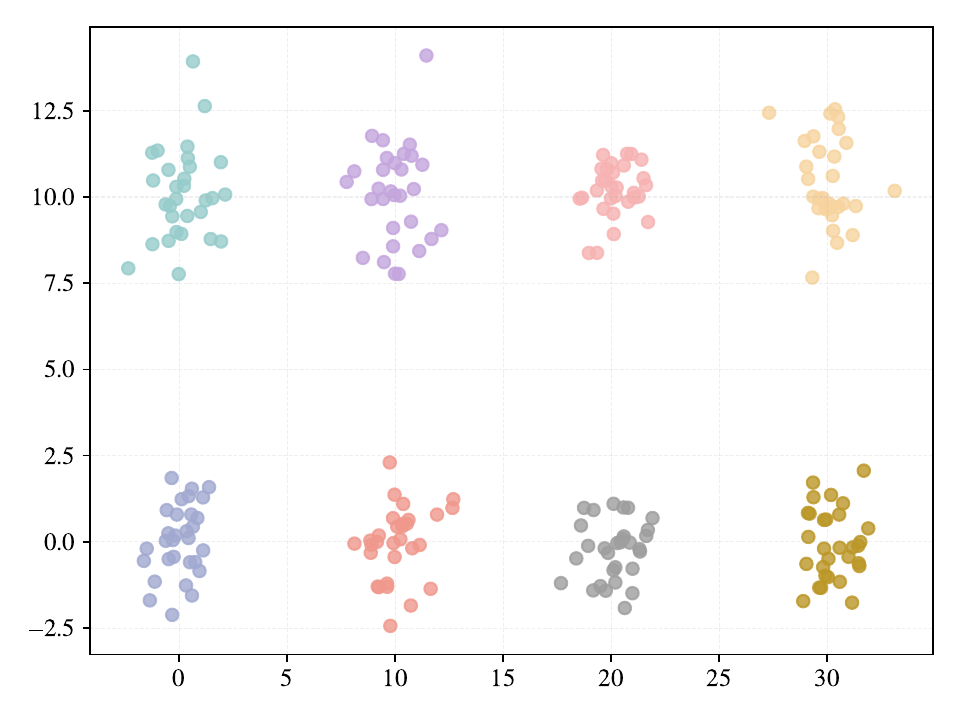}
    \caption{Synthetic dataset used for the expresiveness ablation study.}
    \label{fig:synthetic-points}
\end{figure}

We visualize the synthetic dataset we use for the expressiveness ablation study in Figure \ref{fig:synthetic-points}.

\section{Hyperparameters}

\input{tables/glue_hyper}
\input{tables/cv_hyper}
\input{tables/llama_hyper}

\section{Singular Value Comparison}

\begin{figure}[htb]
  \centering
  \includegraphics[width=\linewidth]{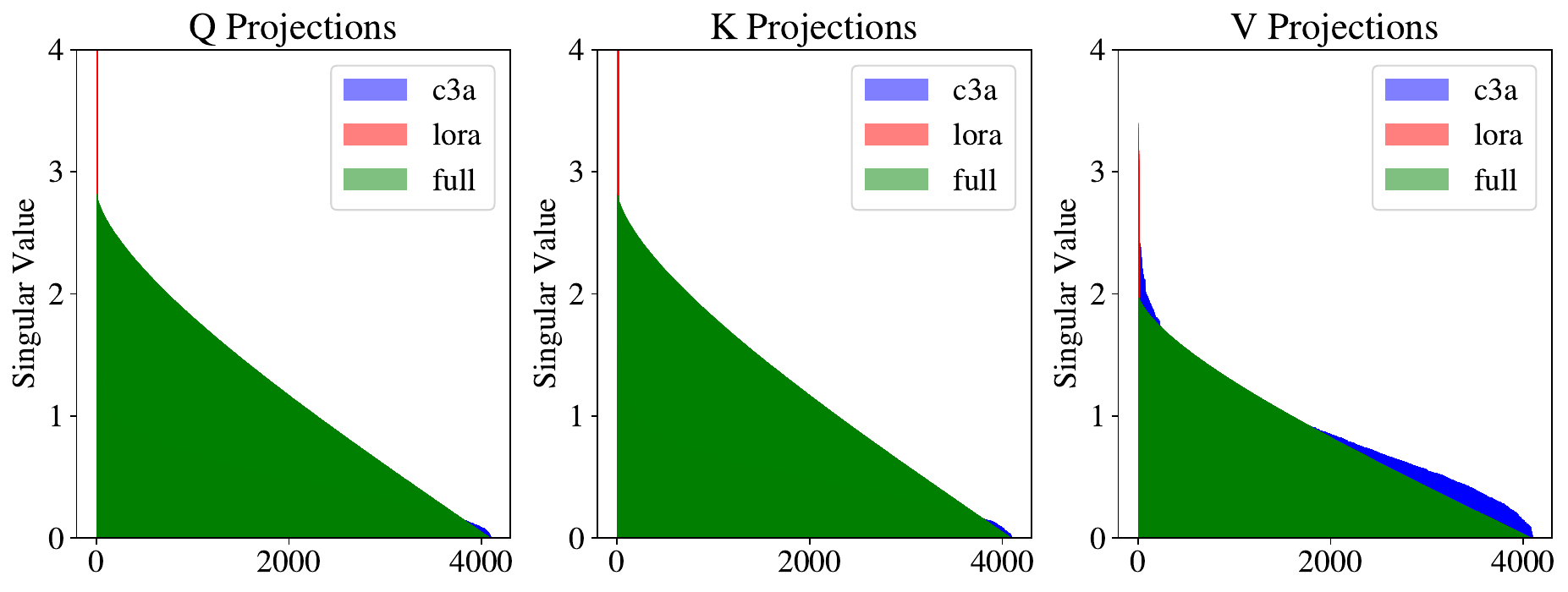}
  \caption{Singular values spectra of the weight difference between C$^3$A, LoRA and Full finetuning.}
  \label{fig:singular_value_comparison}
\end{figure}

%% file: algs/block_circular_convolution.tex
\begin{minipage}{\linewidth}
\begin{algorithm}[H]
\caption{Block-Circular Convolution PyTorch Implementation}
\label{alg:blk-circular-conv}
\lstset{
  basicstyle=\fontsize{7.2pt}{7.2pt}\tt,
  commentstyle=\fontsize{2.2pt}{2.2pt}\color{magenta},
  keywordstyle=\fontsize{7.2pt}{7.2pt}\color{blue},
}

\begin{lstlisting}[language=python]
import torch
from torch.autograd import Function
from torch.fft import fft, ifft

class BlockCircularConvolution(Function):
    @staticmethod
    def forward(ctx, x, w):
        m, n, b = w.shape
        x = x.reshape(*x.shape[:-1], n, b)
        ctx.save_for_backward(x, w)
        x = torch.einsum(
            "...nb,mnb->...mb", ifft(x), fft(w)
        ) 
        x = fft(x).real
        x = x.reshape(*x.shape[:-2], -1)
        return x

    @staticmethod
    def backward(ctx, grad_output):
        x, w = ctx.saved_tensors
        m, n, b = w.shape
        grad_output = grad_output.reshape(
            *grad_output.shape[:-1], m, b
        )
        grad_output_fft = fft(grad_output)
        x_grad = fft(torch.einsum(
            "...mb,mnb->...nb",
            grad_output_fft, ifft(w)
        )).real
        x_grad = x_grad.reshape(
            *x_grad.shape[:-2], -1
        )
        w_grad = fft(torch.einsum(
            "...mb,...nb->mnb",
            grad_output_fft, ifft(x)
        )).real
        return x_grad, w_grad
\end{lstlisting}
\end{algorithm}
\end{minipage}

%% file: algs/fast_delta.tex
\begin{minipage}{\linewidth}
\begin{algorithm}[H]
\caption{Fast Algorithm of Getting $\Delta\mathbf{W}$}
\label{alg:fast-delta}
\lstset{
  basicstyle=\fontsize{7.2pt}{7.2pt}\tt,
  commentstyle=\fontsize{2.2pt}{2.2pt}\color{magenta},
  keywordstyle=\fontsize{7.2pt}{7.2pt}\color{blue},
}

\begin{lstlisting}[language=python]
import torch
from torch.fft import fft, ifft

def get_circulant_fast(w):
    m, n, b = w.shape
    x = torch.eye(n*b)
    x = x.reshape(*x.shape[:-1], n, b)
    x = torch.einsum(
        "...nb,mnb->...mb", ifft(x), fft(w)
    ) 
    x = fft(x).real.flatten(start_dim=1).T
    return x
\end{lstlisting}
\end{algorithm}
\end{minipage}

%% file: tables/vit_dataset_detail.tex
\begin{table}[htb]
\centering
\resizebox{\linewidth}{!}{%
\begin{tabular}{l|lrrrc}
\toprule
Dataset & \#Train & \#Validation & \#Test & \#Class \\ \midrule
Pets \citep{pets} & 3,312 & 368 & 3,669 & 37 \\
Cars \citep{cars} & 7,329 & 815 & 8,041 & 196 \\
DTD \citep{dtd}& 4,060 & 452 & 1,128 & 47 \\
EuroSAT \citep{euro} & 16,200 & 5,400 & 5,400 & 10 \\
FGVC \citep{fgvc} & 3,000 & 334 & 3,333 & 100 \\
RESISC \citep{resisc} & 18,900 & 6,300 & 6,300 & 45 \\ \bottomrule
\end{tabular}%
}
\caption{Details about the vision datasets.}
\label{tab:data-cv}
\end{table}

%% file: tables/image_classification.tex
% \usepackage{booktabs}
% \usepackage{multirow}
% \usepackage{graphicx}

\begin{table*}[htb]
\centering
\resizebox{\textwidth}{!}{
\begin{tabular}{llrccccccc}
\toprule
& Method & \# Params & Pets & Cars & DTD & EuroSAT & FGVC & RESISC & Avg. \\ \midrule
\parbox[t]{4mm}{\multirow{4}{*}{\rotatebox[origin=c]{90}{\textsc{Base}}}} 
& Head & - & 90.28\textsubscript{$\pm$0.43} & 25.76\textsubscript{$\pm$0.28} & 69.77\textsubscript{$\pm$0.67} & 88.72\textsubscript{$\pm$0.13} & 17.44\textsubscript{$\pm$0.43} & 74.22\textsubscript{$\pm$0.10} & 61.03\\
& Full & 85.8M & 92.82\textsubscript{$\pm$0.54} & 85.10\textsubscript{$\pm$0.21} & 80.11\textsubscript{$\pm$0.56} & 99.11\textsubscript{$\pm$0.07} & 61.60\textsubscript{$\pm$1.00} & 96.00\textsubscript{$\pm$0.23} & 85.79\\
\cmidrule(lr){2-10}
& LoRA$_{r=16}$ & 0.59M & 93.76\textsubscript{$\pm$0.44} & 78.04\textsubscript{$\pm$0.33} & 78.56\textsubscript{$\pm$0.62} & 98.84\textsubscript{$\pm$0.08} & \textbf{56.64}\textsubscript{$\pm$0.55} & \textbf{94.66}\textsubscript{$\pm$0.17} & 83.42\\
& C$^3$A$_{b=768/12}$ & 0.22M & \textbf{93.88}\textsubscript{$\pm$0.22} & \textbf{79.05}\textsubscript{$\pm$0.35} & \textbf{80.57}\textsubscript{$\pm$0.53} & \textbf{98.88}\textsubscript{$\pm$0.07} & 54.31\textsubscript{$\pm$0.79} & 94.54\textsubscript{$\pm$0.23} & \textbf{83.54}\\

\midrule[0.6pt]
\midrule

\parbox[t]{4mm}{\multirow{4}{*}{\rotatebox[origin=c]{90}{\textsc{Large}}}} 
& Head & - & 91.11\textsubscript{$\pm$0.30} & 37.91\textsubscript{$\pm$0.27} & 73.33\textsubscript{$\pm$0.26} & 92.64\textsubscript{$\pm$0.08} & 24.62\textsubscript{$\pm$0.24} & 82.02\textsubscript{$\pm$0.11} & 66.94\\
& Full & 303M & 94.30\textsubscript{$\pm$0.31} & 88.15\textsubscript{$\pm$0.50} & 80.18\textsubscript{$\pm$0.66} & 99.06\textsubscript{$\pm$0.10} & 67.38\textsubscript{$\pm$1.06} & 96.08\textsubscript{$\pm$0.20} & 87.53\\
\cmidrule(lr){2-10}
& LoRA$_{r=16}$ & 1.57M & \textbf{94.62}\textsubscript{$\pm$0.47} & \textbf{86.11}\textsubscript{$\pm$0.42} & 80.09\textsubscript{$\pm$0.42} & \textbf{98.99}\textsubscript{$\pm$0.03} & 63.64\textsubscript{$\pm$0.83} & 95.52\textsubscript{$\pm$0.21} & 86.56 \\
& C$^3$A$_{b=1024/16}$ & 0.79M & 94.48\textsubscript{$\pm$0.30} & 84.94\textsubscript{$\pm$0.39} & \textbf{82.62}\textsubscript{$\pm$0.52} & 98.75\textsubscript{$\pm$0.19} & \textbf{63.80}\textsubscript{$\pm$0.37} & \textbf{95.94}\textsubscript{$\pm$0.16} & \textbf{86.69} \\
\bottomrule
\end{tabular}%
}
\caption{Fine-tuning results with ViT-Base and ViT-Large models on various image classification datasets. The models are fine-tuned for 10 epochs, and the best-performing model, based on validation set accuracy, is selected. The reported accuracy corresponds to the performance on the test set. The best results between LoRA and C$^3$A for each dataset are highlighted in \textbf{bold}. ``Avg.'' denotes the average accuracy of each method across all datasets.}
\label{tab:image_classification}
\end{table*}

%% file: tables/glue_details.tex
\begin{table*}[htb]
\centering
\resizebox{1.\textwidth}{!}{%
\begin{tabular}{@{}llrrrrll@{}}
\toprule
\multicolumn{1}{l}{\textbf{Corpus}} & Task & \# Train & \# Val & \# Test & \# Labels & Metrics & Domain \\ \midrule
\multicolumn{8}{c}{Single-Sentence Tasks} \\ \midrule
\multicolumn{1}{l}{CoLA} & Acceptability & 8.55k & 1.04k & 1.06k & 2 & Matthews Corr. & misc. \\
\multicolumn{1}{l}{SST-2} & Sentiment & 67.3k & 872 & 1.82k & 2 & Accuracy & Movie reviews \\ \midrule
\multicolumn{8}{c}{Similarity and Paraphrase Tasks} \\ \midrule
\multicolumn{1}{l}{MRPC} & Paraphrase & 3.67 & 408 & 1.73k & 2 & Accuracy/F1 & News \\
\multicolumn{1}{l}{STS-B} & Sentence similarity & 5.75k & 1.5k & 1.38k & 1 & Pearson/Spearman Corr. & misc. \\
\multicolumn{1}{l}{QQP} & Paraphrase & 364k & 40.4k & 391k & 2 & Accuracy/F1 & Social QA \\ \midrule
\multicolumn{8}{c}{Inference Tasks} \\ \midrule
\multicolumn{1}{l}{MNLI} & NLI & 393k & 19.65k & 19.65k & 3 & Accuracy & misc. \\
\multicolumn{1}{l}{QNLI} & QA/NLI & 105k & 5.46k & 5.46k & 2 & Accuracy & Wikipedia \\
\multicolumn{1}{l}{RTE} & NLI & 2.49k & 277 & 3k & 2 & Accuracy & News \& Wikipedia \\ \bottomrule
\end{tabular}%
}
\caption{Task descriptions and dataset statistics of the GLUE benchmark \citep{wang2018glue}.}
\label{tab:data-glue}
\end{table*}

%% file: tables/glue_hyper.tex
% \usepackage{booktabs}
% \usepackage{multirow}
% \usepackage{graphicx}
\begin{table*}[htb]
\centering
\begin{tabular}{@{}cl|cccccc@{}}
\toprule
Model & Hyperparameter & \multicolumn{1}{c}{SST-2} & \multicolumn{1}{c}{MRPC} & \multicolumn{1}{c}{CoLA} & \multicolumn{1}{c}{QNLI} & \multicolumn{1}{c}{RTE} & \multicolumn{1}{c}{STS-B} \\ \midrule
\multirow{5}{*}{\rotatebox{90}{Both}} & Optimizer & \multicolumn{6}{c}{AdamW} \\
 & LR Schedule & \multicolumn{6}{c}{Linear} \\
 & Warmup Ratio & \multicolumn{6}{c}{0.06} \\
 & C$^3$A Initialization & \multicolumn{6}{c}{Xavier Uniform} \\
 & Max Seq. Len & \multicolumn{6}{c}{512} \\
 \midrule
\multirow{4}{*}{\rotatebox{90}{Base}} & Epochs & 40 & 80 & 80 & 40 & 80 & 80\\
 & Batch Size & 128 & 128 & 128 & 64 & 64 & 128 \\
 & Learning Rate (C$^3$A$_{b=768/6}$) & 2E-1 & 3E-1 & 2E-1 & 7E-2 & 3E-1 & 2E-1 \\
 & Learning Rate (Head) & 2E-4 & 4E-6 & 3E-2 & 8E-6 & 6E-3 & 4E-2 \\
 \midrule
\multicolumn{1}{c}{\multirow{4}{*}{\rotatebox{90}{Large}}} & Epochs & 10 & 80 & 70 & 30 & 60 & 40 \\
\multicolumn{1}{l}{} & Batch Size & 128 & 128 & 128 & 32 & 64 & 128 \\
\multicolumn{1}{l}{} & Learning Rate (C$^3$A$_{b=1024/8}$) & 9E-2 & 3E-1 & 2E-1 & 7E-2 & 5E-2 & 2E-1 \\
\multicolumn{1}{l}{} & Learning Rate (Head) & 2E-4 & 5E-6 & 3E-3 & 8E-6 & 3E-3 & 5E-4 \\ \midrule
\end{tabular}%
\caption{Hyperparameter setup of C$^3$A for the GLUE benchmark.}
\label{tab:hyper-glue}
\end{table*}

%% file: tables/cv_hyper.tex
% \usepackage{booktabs}
% \usepackage{graphicx}
\begin{table*}[htb]
\centering
\begin{tabular}{@{}cl|cccccc@{}}
\toprule
Model & Hyperparameter & \multicolumn{1}{c}{Pets} & \multicolumn{1}{c}{Cars} & \multicolumn{1}{c}{DTD} & \multicolumn{1}{c}{EuroSAT} & \multicolumn{1}{c}{FGVC} & \multicolumn{1}{c}{RESISC} \\ \midrule
\multirow{5}{*}{\rotatebox{90}{Both}} & Optimizer & \multicolumn{6}{c}{AdamW} \\
 & LR Schedule & \multicolumn{6}{c}{None} \\
 & C$^3$A Initialization & \multicolumn{6}{c}{Xavier Uniform} \\
 & Epochs & \multicolumn{6}{c}{10} \\
 & Batch Size & \multicolumn{6}{c}{64} \\
 \midrule
 \multirow{3}{*}{\rotatebox{90}{Base}} & Learning Rate (C$^3$A$_{b=768/12}$) & 4E-1 & 4E+0 & 2E+0 & 2E+0 & 7E+0 & 2E+0 \\
 & Learning Rate (Head) & 1E-2 & 1E-2 & 2E-2 & 8E-3 & 1E-2 & 2E-2 \\
  & Weight Decay & 3E-4 & 5E-4 & 6E-5 & 2E-5 & 1E-5 & 2E-5 \\
 \midrule
 \multirow{3}{*}{\rotatebox{90}{Large}}
 & Learning Rate (C$^3$A$_{b=1024/16}$) & 7E-1 & 4E+0 & 2E+0 & 2E+0 & 4E+0 & 3E+0 \\
 & Learning Rate (Head) & 3E-3 & 8E-3 & 7E-3 & 2E-2 & 1E-1 & 4E-3 \\ 
 & Weight Decay & 4E-3 & 1E-5 & 2E-4 & 5E-4 & 2E-5 & 9E-5 \\ \midrule
\end{tabular}%
\caption{Hyperparameter setup of C$^3$A for image classification tasks.}
\label{tab:hyper-cv}
\end{table*}

%% file: tables/llama_hyper.tex
\begin{table}[htb]
    \centering

    \resizebox{\linewidth}{!}{
        \begin{tabular}{l|*{3}{c}}
            \toprule
            Hyperparameter & Common & Math & Code \\
            \midrule
            Optimizer & \multicolumn{3}{c}{AdamW} \\
            LR Scheduler & \multicolumn{3}{c}{Cosine} \\
            Batch Size & 16 & 128 & 128 \\
            Warmup Ratio & 0.05 & 0.03 & 0.03 \\
            Epoch & 3 & 2 & 2 \\
            Dropout & \multicolumn{3}{c}{0.05} \\
            \midrule
            % \multirow{1}{*}{LLaMA2-7B}
            % & Learning Rate (LoRA) & 5E-4 & 5E-4 & 5E-4 \\
            % & Learning Rate (DoRA) & 4E-4 & 5E-4 & 4E-4 \\
            Learning Rate (LLaMA2-7B) & 3E-1 & 4E-1 & 6E-1 \\
            
            % \midrule
            % \multirow{1}{*}{LLaMA3-8B}
            % & Learning Rate (LoRA) & 5E-4 & 5E-4 & 4E-4 & 5E-4 & 4E-4 & 4E-4 & 5E-4 \\
            % & Learning Rate (DoRA) & 6E-4 & 2E-4 & 5E-4 & 5E-4 & 4E-4 & 4E-4 & 4E-4 \\
            Learning Rate (LLaMA3-8B) & 3E-1 & 5E-1 & 6E-1 \\
            \bottomrule
        \end{tabular}
    }
    \caption{Hyperparameter setup of C$^{3}$A for instruction tuning.}
    \label{tab:llama_hyperparameters}
\end{table}